# A Novel Feedforward Youla Parameterization Method for Avoiding Local Minima in Stereo Image Based Visual Servoing Control

**Rongfei Li** [1] **and Francis Assadian** [2]

[1] University of California, Davis; rfli@ucdavis.edu
[2] University of California, Davis; fassadian@ucdavis.edu

**Abstract**: In robot navigation and manipulation, accurately determining the camera's pose relative to the environment is crucial for effective task execution. In this paper, we systematically prove that this problem corresponds to the Perspective-3-Point (P3P) formulation, where exactly three known 3D points and their corresponding 2D image projections are used to estimate the pose of a stereo camera. In image-based visual servoing (IBVS) control, the system becomes overdetermined, as the 6 degrees of freedom (DoF) of the stereo camera must align with 9 observed 2D features in the scene. When more constraints are imposed than available DoFs, global stability cannot be guaranteed, as the camera may become trapped in a local minimum far from the desired configuration during servoing. To address this issue, we propose a novel control strategy for accurately positioning a calibrated stereo camera. Our approach integrates a feedforward controller with a Youla parameterization-based feedback controller, ensuring robust servoing performance. Through simulations, we demonstrate that our method effectively avoids local minima and enables the camera to reach the desired pose accurately and efficiently.

## 1. Introduction

Determining the accurate pose of the camera is a fundamental problem in robot manipulations, as it provides the spatial transformation needed to map 3D world points to 2D image coordinates. The task involving camera pose estimation is essential for various applications, such as augmented reality [1], 3D reconstruction [2], SLAM [3], and autonomous navigation [4]. This becomes especially critical when robots operate in unstructured, fast-changing, and dynamic environments, performing tasks such as human-robot interaction, accident recognition and avoidance, and eye-in-hand visual servoing. In such scenarios, accurate camera pose estimation ensures that visual data is readily available for effective robotic control [5].

A classic approach to estimating the pose of a calibrated camera is solving the Perspective-n-Point (PnP) problem [6], which establishes a mathematical relationship between a set of n 3D points in the world and their corresponding 2D projections in an image. To uniquely determine the pose of a monocular camera in space, it is a Perspective-4-Point (P4P) problem, where exactly 4 known 3D points and their corresponding 2D image projections are used. Bujnak et al. [7] generalize four solutions for P3P problem while giving a single unique solution existed for P4P problem in a fully calibrated camera scenario. To increase accuracy, modern PnP approaches considers more than three 2D-3D correspondences. Among PnP solutions, EPnP (Efficient PnP) method finds the optimal estimation of pose from a linear system that expresses each reference point as a weighted sum of four virtual control points [8]. Another advanced approach, SQPnP (Sparse

Quadratic PnP) formulates the problem as a sparse quadratic optimization, achieving enhanced accuracy by minimizing a sparse cost function [9].

In recent years, many other methods have been developed to show improved accuracy than PnP based methods. For instance, Alkhatib et al. [10] utilize Structure from Motion (SfM) to estimate a camera's pose by extracting and matching key features across various images taken from different viewpoints to establish correspondences. Moreover, Wang et al. [11] introduce visual odometry into camera's pose estimations based on the movement between consecutive frames. In addition, recent advancements in deep learning have led to the development of models, such as Convolutional Neural Networks, specifically tailored for camera pose estimation [12-14]. However, these advanced methods often come with significant computational costs, requiring multiple images from different perspectives for accurate estimation. In contrast, PnP-based approaches offer a balance between accuracy and efficiency, as they can estimate camera pose from a single image, making them highly suitable for real-time applications such as navigation and scene understanding.

In image-based visual servoing (IBVS) [15], the primary goal is to control a robot's motion using visual feedback. Accurate real-time camera pose estimation is crucial for making informed control decisions, particularly in eye-in-hand (EIH) configurations [16-17], where a camera is mounted directly on a robot manipulator. In this setup, robot motion directly induces camera motion, making precise pose estimation essential. Due to its computational efficiency, PnP-based approaches remain widely applied in real-world IBVS tasks [18-20]. The PnP process begins by establishing correspondences between 3D feature points and their 2D projections in the camera image. The PnP algorithm then computes the camera pose from these correspondences, translating the geometric relationship into a format that the IBVS controller can use. By detecting spatial discrepancies between the current and desired camera poses, the robot can adjust its movements accordingly.

However, PnP-based IBVS presents challenges for visual control in robotics. One key issue is that IBVS often results in an overdetermined system, where the number of visual features exceeds the number of joint variables available for adjustment. For example, at least four 2D-3D correspondences are needed for a unique pose solution [6], but a camera's full six-degree-of-freedom (6-DOF) pose means that a 6-DOF robot may need to align itself with eight or more observed features. In traditional IBVS [15], the interaction matrix (or image Jacobian) defines the relationship between feature changes and joint velocities. When the system is overdetermined, this matrix contains more constraints than joint variables, leading to redundant information. Research [15] suggests that this redundancy may cause the camera to converge to local minima, failing to reach the desired pose. Although local asymptotic stability is always ensured in IBVS, global asymptotic stability cannot be guaranteed when the system is overdetermined.

In recent research on visual servoing, various methods have been developed to address the limitations of traditional Image-Based Visual Servoing (IBVS), particularly the issue of local minima due to system overdetermination. Gans et al. [21] proposed a hybrid switched-system control system, which alternates between IBVS and Position-Based Visual Servoing (PBVS) modes based on Lyapunov functions. This method offers formal local stability and effectively prevents loss of visual features or divergence from the goal. However, it may suffer from discontinuities near switching surfaces and is limited by local convergence guarantees. Chaumette et al. [22] developed a 2½ D visual servoing method, which integrates both image and pose features into a block-triangular Jacobian structure, allowing for smoother and more predictable camera trajectories while avoiding singularities. Yet, it is sensitive to depth estimation errors and requires careful selection of hybrid features. Roque et al. [23] introduced an MPC-based IBVS framework for quadrotors, which leverages model predictive control to handle under-actuation while tracking IBVS commands. It provides formal stability guarantees and supports constraint handling in real time, but its reliance on linearization around hover and the absence of direct coupling with visual dynamics limit its generality and precision in highly nonlinear settings.

While these methods achieve significant improvements in most scenarios, they also introduce computational challenges compared to traditional IBVS. The switched control method requires different control strategies tailored to specific dynamics, increasing the complexity of the overall control architecture [24]. The 2-1/2-D visual servoing method demands real-time processing of both visual and positional data, which can impose significant computational loads and limit performance in dynamic environments [25]. MPC approaches introduce additional computational overhead by requiring complex optimization at every time step, making real-time implementation costly [26].

In this paper, we focus on the PnP framework for determining and controlling the pose of a stereo camera within an image-based visual servoing (IBVS) architecture. In traditional IBVS, depth information between objects and the image plane is crucial for developing the interaction matrix. However, with a monocular camera, depth can only be estimated or approximated using various algorithms [15], and inaccurate depth estimation may lead to system instability. In contrast, a stereo camera system can directly measure depth through disparity between two image planes, enhancing system stability.

A key novelty of this paper is providing a systematic proof that stereo camera pose determination in IBVS can be formulated as a P3P (Perspective-3-Point) problem, which, to the best of our knowledge, has not been explored in previous research. Since three corresponding points, totaling nine coordinates, are used to control the six DoFs camera pose, the IBVS control system for a stereo camera is overdetermined, leading to the potential issue of local minima during control maneuvers. While existing approaches can effectively address local minima, they often introduce excessive computational overhead, making them impractical for high-speed real-world applications.

To address this challenge, we propose a feedforward-feedback control architecture. The feedback component follows a cascaded control loop based on the traditional IBVS framework [15], where the inner loop handles robot joint rotation, and the outer loop generates joint angle targets based on visual data. One key improvement in this work is the incorporation of both kinematics and dynamics during the model development stage. Enhancing model fidelity in the control design improves pose estimation precision and enhances system stability, particularly for high-speed tasks. Both control loops are designed using Youla parameterization [27], a robust control technique that enhances resistance to external disturbances. The feedforward controller takes target joint configurations, which are associated with the desired camera pose as inputs, ensuring a fast system response while avoiding local minima traps. Simulation results presented in this paper demonstrate that the proposed control system effectively moves the stereo camera to its desired pose accurately and efficiently, making it well-suited for high-speed robotic applications.

## 2. System Configuration

An eye-in-hand robotic system has been developed to precisely control the pose of a stereo camera system, as illustrated in Figure 1. The robotic manipulator is equipped with six revolute joints, allowing unrestricted movement of the camera across six degrees of freedom (DoFs)—three for positioning and three for orientation. Assume a set of fiducial markers is placed in the workspace, with their coordinates fixed and predefined in an inertial frame. Utilizing the Hough transform [28] in computer vision, these markers can be detected and localized by identifying their centers in images captured by the stereo camera system. The control system within the robotic manipulator aligns the camera to its desired pose by matching the detected 2D features in the current frame with target 2D features. Throughout this process, it is assumed that all fiducial markers remain within the camera's field of view. As depicted in Figure 1, multiple Cartesian coordinate systems are illustrated. The base frame {O} serves as an inertial reference fixed to the bottom of the robot manipulator, while the camera frame {C} is a body-fixed frame attached to the robot's end-effector.

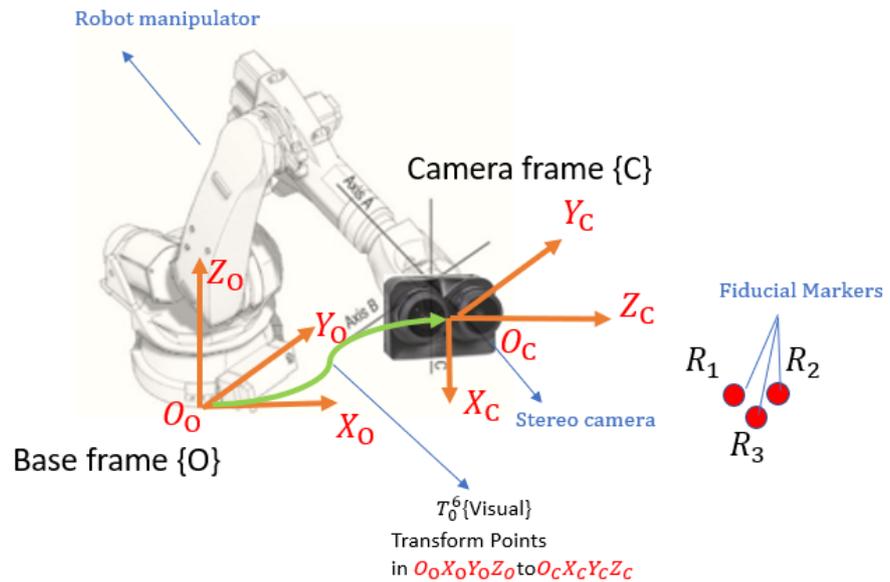

**Figure 1.** The eye-in-hand robot configuration.

## 3. Proof of P3P for the Stereo Camera System

Given its intrinsic parameters and a set of n correspondences between its 3D points and 2D projections to determine the camera's pose is known as perspective-n-point (PnP) problem. This well-known work [7] has proved that at least four correspondences are required to uniquely determine the pose of a monocular camera, a situation referred to as the P4P problem.

To illustrate, consider the P3P case for a monocular camera. Let points A, B, and C exist in space, with $O_1$, $O_2$ and $O_3$ representing different perspective centers. The angles ∠ AOB, ∠ AOC and ∠ BOC remain the same across all three perspectives. Given a fixed focal length, the image coordinates of points A, B, and C will be identical when observed from these three perspectives. In other words, it is impossible to uniquely identify the camera's pose based solely on the image coordinates of three points.

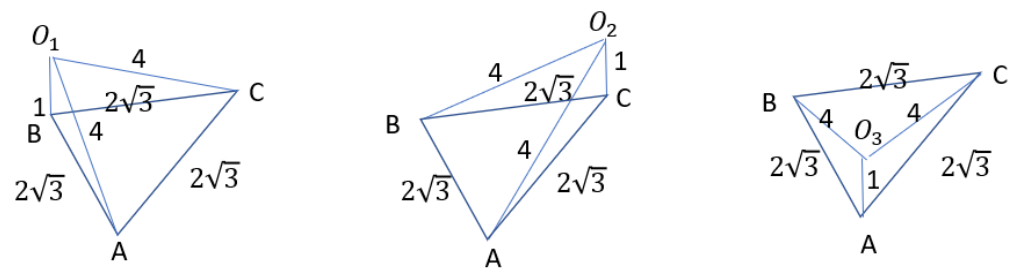

**Figure 2.** P3P case of a monocular camera.

The PnP problem with a stereo camera has not been thoroughly addressed in prior research. A stereo camera can detect three image coordinates of a 3D point in space. This paper proposes that a complete solution to the PnP problem for a stereo camera can be framed as a P3P problem. Below is the complete proof of this proposition.

*Proof:*

For a stereo system, if all intrinsic parameters are fixed and given, we can readily compute the 3D coordinates of an object point given the image coordinates of that point.

This provides a unique mapping from the image coordinates of a point to its corresponding 3D coordinates in a Cartesian frame. The orientation and position of the camera system uniquely define the origin and axis orientations of this Cartesian coordinate system in space. Consequently, PnP problem can be framed as follows: given n points with their 3D coordinates measured in an unknown Cartesian coordinate system in space, what is the minimum number n required to accurately determine the position and orientation of the 3D Cartesian coordinate frame established in that space?

### 1) P1P problem with the stereo camera:

If we know the coordinates of a single point in space, defined by a 3D Cartesian coordinate system, an infinite number of corresponding coordinate systems can be established. Any such coordinate system can have its origin placed on the surface of a sphere centered at this point, with a radius $R = \sqrt{X^{C^2} + Y^{C^2} + Z^{C^2}}$, where $[X^C, Y^C, Z^C]^T$ are the coordinates measured by the Cartesian system (see Figure 3).

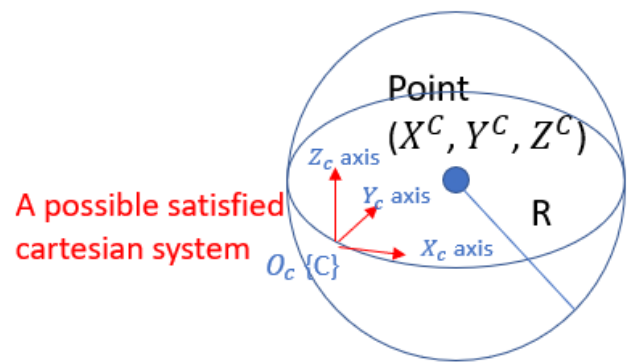

**Figure 3.** P1P Problem with a Stereo Camera System

### 2) P2P problem with the stereo camera:

When the coordinates of two points in space are known, an infinite number of corresponding coordinate systems can be established. Any valid coordinate system can have its origin positioned on a circle centered at point $O$ with a radius $R$ as illustrated in Figure 4. This circle is constrained by the triangle formed by points $O_c$, $P_1$, and $P_2$, where the sides of the triangle are defined by the lengths $R_1, R_2$ and $D$. Specifically, $R_1 = \sqrt{X_1^{C^2} + Y_1^{C^2} + Z_1^{C^2}}$, $R_2 = \sqrt{X_2^{C^2} + Y_2^{C^2} + Z_2^{C^2}}$, $D = \sqrt{(X_1^C - X_2^C)^2 + (Y_1^C - Y_2^C)^2 + (Z_1^C - Z_2^C)^2}$. The radius of the circle $R$ corresponds to the height of the base $D$ of the triangle. The center of the circle $O$ is located at the intersection of the height and the base.

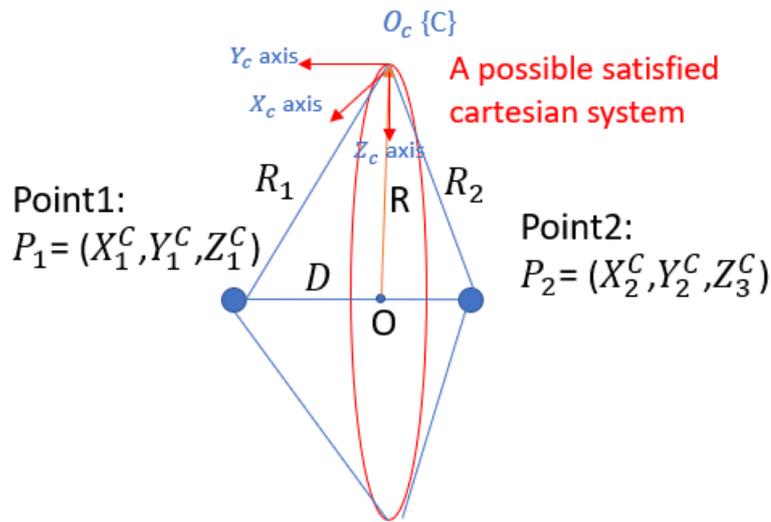

**Figure 4.** P2P problem with a stereo camera system. *All potential cartesian systems are located on the circle plotted in red.*

### 3) P3P problem with the stereo camera:

When three points in space are known, and the lines connecting these points are not collinear, we can uniquely establish one coordinate system. As illustrated in the figure below, three non-collinear points define a plane in space, which has a uniquely defined normal unit vector $\vec{n}$. Given the coordinates of the three points, we can calculate vectors as follows: the vector $\vec{P_1P_2} = (X_2^C - X_1^C, Y_2^C - Y_1^C, Z_2^C - Z_1^C)$, and the vector $\vec{P_1P_3} = (X_3^C - X_1^C, Y_3^C - Y_1^C, Z_3^C - Z_1^C)$. The unit vector $\vec{n}$ which is perpendicular to the plane formed by these three points, can be expressed as:

$$\vec{n} = \frac{\vec{P_1P_2} \times \vec{P_1P_3}}{|\vec{P_1P_2} \times \vec{P_1P_3}|} \tag{1}$$

Here, × denotes the cross product.

The angles between $\vec{n}$ and XYZ axes of the coordinate system can be expressed as follows:

$$\cos(\theta_x) = \vec{n} \cdot \vec{\imath} \tag{2}$$

$$\cos(\theta_y) = \vec{n} \cdot \vec{\jmath} \tag{3}$$

$$\cos(\theta_z) = \vec{n} \cdot \vec{k} \tag{4}$$

Where $\theta_x, \theta_y$ and $\theta_z$ are the angles between $\vec{n}$ and the unit vectors in the X, Y, and Z directions, denoted as $\vec{\imath}, \vec{\jmath},$ and $\vec{k}$ respectively. Therefore, with the direction $\vec{n}$ fixed in space, the orientations of each axis of the coordinate system can be computed uniquely.

According to the P1P problem, the origin of the coordinate system must lie on the surface of a sphere centered at $P_1$ with radius $= \sqrt{X_1^{C2} + Y_1^{C2} + Z_1^{C2}}$ as depicted in Figure 3. Each coordinate system established with a different origin point on the surface of this sphere results in a unique configuration of the axis orientations. Therefore, as the orientations of the axes are defined in space, the position of the frame (or the position of the origin) is also uniquely defined.

In conclusion, the P3P problem is sufficient to solve the PnP problem for a stereo camera system.

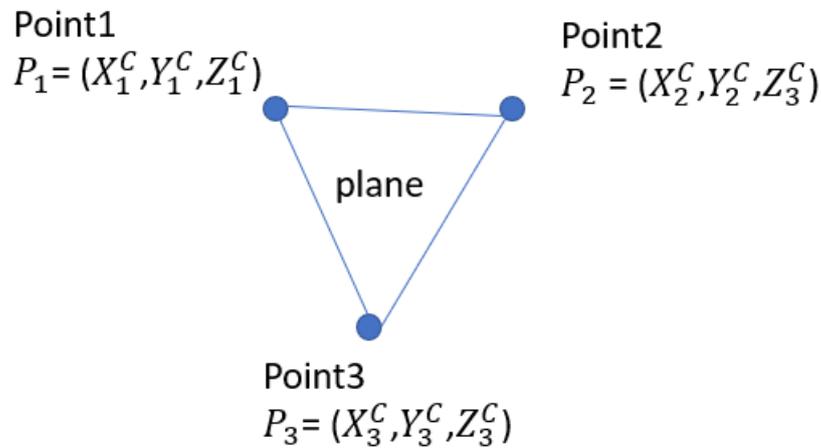

Figure 5. P3P Problem with a Stereo Camera System.

***Prove concluded***

This proposition indicates that to uniquely determine the full 6 DoFs of the stereo camera, at least three points (or nine 2D features) are required to match in the image-based visual servoing control.

## 4. Model Development

### 4.1. Stereo Camera Model

Depth between the objects to the camera plane is either approximated or estimated in the IBVS for generating the interaction matrix [15]. Using a stereo camera system in IBVS eliminates the inaccuracies associated with monocular depth estimation, as it directly measures depth by leveraging the disparity between the left and right images.

The stereo camera model is illustrated in Figure 6. A stereo camera consists of two lenses separated by a fixed baseline $b$. Each lens has a focal length $F$ (measured in mm) which is the distance from the image plane to the focal point. Assuming the camera is calibrated, the intrinsic parameters: $b$, $F$ is accurately estimated. A scene point $I$ is measured in the 3D coordinate frame {C} centered at the middle of the baseline with its coordinates as $[X^C, Y^C, Z^C]^T$. The stereo camera model maps the 3D coordinates of this point to its 2D coordinates projected on the left and right image plane as $[u_l, v]^T$ and $[u_r, v]^T$, respectively. The full camera projection map, incorporating both intrinsic and extrinsic parameters, is given by:

$$s \cdot p_{image} = K \cdot [R|T] \cdot P_C \qquad (5)$$

Where $p_{image}$ are the image coordinates of the point and $P_C$ are the 3D coordinates measured in the camera frame {C}. $s$ is the scale factor that ensures correct projection between 2D and 3D features. $K$ is the intrinsic matrix with a size of 3X3, and the mathematical expression is presented as:

$$K = \begin{bmatrix} F & k & u_0 \\ 0 & F & v_0 \\ 0 & 0 & 1 \end{bmatrix} \qquad (6)$$

Where $k$ is the skew factor, which represents the angle between the image axes ($u$ and $v$ axis). $u_0$ and $v_0$ are coordinates offsets in image planes.

In equation (5), $R$ is the rotational matrix from camera frame {C} to each image coordinate frame, and $T$ is the translation matrix from camera frame {C} to each camera lens center. Since there is no rotation between the camera frame {C} and image frames but only a translation along the $X_C$ axis occurs, the transformation matrices for the left and right image planes are expressed as:

$$[R|T]_{Left} = \begin{bmatrix} 1 & 0 & 0 & -b/2 \\ 0 & 1 & 0 & | & 0 \\ 0 & 0 & 1 & 0 \end{bmatrix} \tag{7}$$

$$[R|T]_{Right} = \begin{bmatrix} 1 & 0 & 0 & b/2 \\ 0 & 1 & 0 & | & 0 \\ 0 & 0 & 1 & 0 \end{bmatrix} \tag{8}$$

Assume the u and v axis are perfectly perpendicular (take k = 0), and there are no offsets in the image coordinates (take $u_0 = v_0 = 0$ ) for both lens. Also, set factor $s = Z_I^C$ accounts for perspective depth scaling. The projection equations for the left and right image planes can be rewritten in homogeneous coordinates as:

$$Z_I^C \cdot \begin{bmatrix} u_l \\ v \\ 1 \end{bmatrix} = \begin{bmatrix} F & 0 & 0 \\ 0 & F & 0 \\ 0 & 0 & 1 \end{bmatrix} \cdot \begin{bmatrix} 1 & 0 & 0 & -b/2 \\ 0 & 1 & 0 & | & 0 \\ 0 & 0 & 1 & 0 \end{bmatrix} \cdot \begin{bmatrix} X_I^C \\ Y_I^C \\ Z_I^C \\ 1 \end{bmatrix} \tag{9}$$

$$Z_I^C \cdot \begin{bmatrix} u_r \\ v \\ 1 \end{bmatrix} = \begin{bmatrix} F & 0 & 0 \\ 0 & F & 0 \\ 0 & 0 & 1 \end{bmatrix} \cdot \begin{bmatrix} 1 & 0 & 0 & b/2 \\ 0 & 1 & 0 & | & 0 \\ 0 & 0 & 1 & 0 \end{bmatrix} \cdot \begin{bmatrix} X_I^C \\ Y_I^C \\ Z_I^C \\ 1 \end{bmatrix} \tag{10}$$

Equation (9) and (10) establish the mathematical relationship between the 3D coordinates of a point in the camera frame {C} and its 2D projections on the left and right image planes. The pixel value along the $v$ -axis remains the same for both images. As a result, a scene point's 3D coordinates can be mapped to a set of three image coordinates in the stereo camera system, expressed as:

$$\text{Stereo-camera mapping } M: \ P^C = [X^C, Y^C, Z^C]^T \rightarrow p_{image} = [u_l, u_r, v]^T \tag{11}$$

The mapping function $M$ is nonlinear and depends on the stereo camera parameters $P_{a\_camera}$, specifically $b$, and $F$.

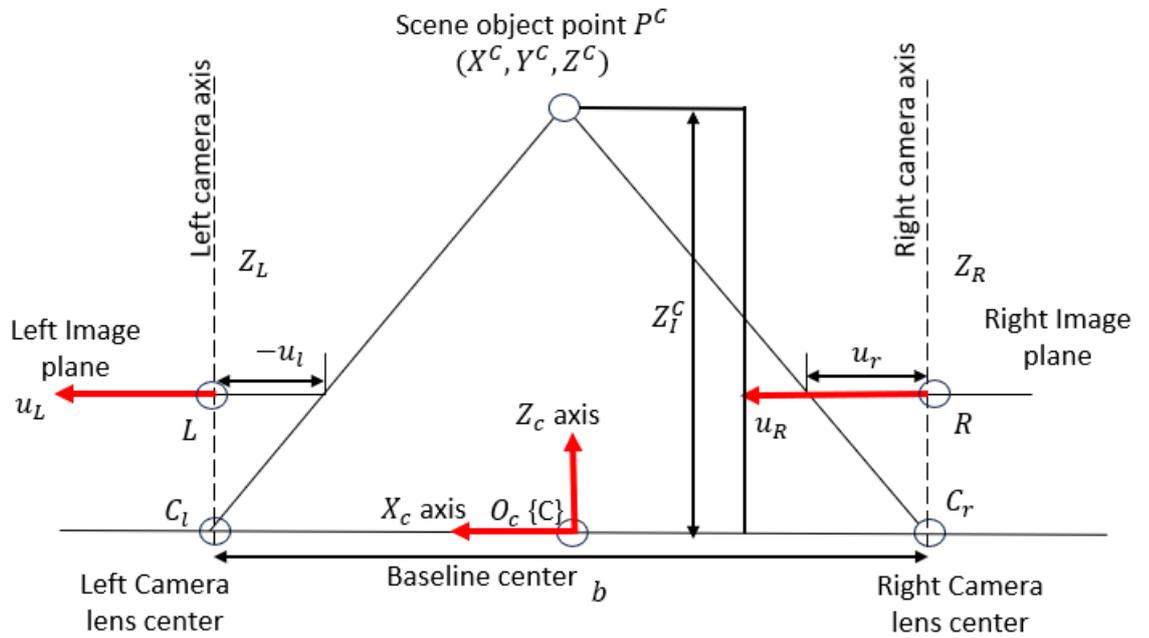

**Figure 6.** The projection of a scene object on the stereo camera's image planes. *Note: The v-coordinate on each image plane is not displayed in this plot but is measured along the axis that is perpendicular to and pointing out of the plot.*

### 4.2. Robot Manipulator Forward Kinematic Model

A widely used method for defining and generating reference frames in robotic applications is the Denavit-Hartenberg (D-H) convention [29]. In this approach, each robotic link is associated with a Cartesian coordinate frame $O_i X_i Y_i Z_i$. According to the D-H convention, the homogeneous transformation matrix $A_i^{i-1}$, which represents the transformation from frame $i-1$ to frame $i$, can be decomposed into a sequence of four fundamental transformations:

$$A_i^{i-1} = Rot_{z,q_i} Trans_{z,d_i} Trans_{x,a_i} Rot_{x,\alpha_i}$$

Expanding the transformation into its matrix form:

$$A_i^{i-1} = \begin{bmatrix} c_{q_i} & -s_{q_i} & 0 & 0 \\ s_{q_i} & c_{q_i} & 0 & 0 \\ 0 & 0 & 1 & 0 \\ 0 & 0 & 0 & 1 \end{bmatrix} \begin{bmatrix} 1 & 0 & 0 & 0 \\ 0 & 1 & 0 & 0 \\ 0 & 0 & 1 & d_i \\ 0 & 0 & 0 & 1 \end{bmatrix} \begin{bmatrix} 1 & 0 & 0 & a_i \\ 0 & 1 & 0 & 0 \\ 0 & 0 & 1 & 0 \\ 0 & 0 & 0 & 1 \end{bmatrix} \begin{bmatrix} 1 & 0 & 0 & 0 \\ 0 & c_{\alpha_i} & -s_{\alpha_i} & 0 \\ 0 & s_{\alpha_i} & c_{\alpha_i} & 0 \\ 0 & 0 & 0 & 1 \end{bmatrix} \quad (12)$$

$$= \begin{bmatrix} c_{q_i} & -s_{q_i} c_{\alpha_i} & s_{q_i} s_{\alpha_i} & a_i c_{q_i} \\ s_{q_i} & c_{q_i} c_{\alpha_i} & -c_{q_i} s_{\alpha_i} & a_i s_{q_i} \\ 0 & s_{\alpha_i} & c_{\alpha_i} & d_i \\ 0 & 0 & 0 & 1 \end{bmatrix} \quad (13)$$

Note: $c_{\theta_i} \equiv \cos(q_i), \ c_{\alpha_i} \equiv \cos(\alpha_i), \ s_{\theta_i} \equiv \sin(q_i), \ s_{\alpha_i} \equiv \sin(\alpha_i)$ \quad (14)

The parameters $q_i$, $a_i$, $\alpha_i$ and $d_i$ define the link and joint characteristics of the robot. Here, $a_i$ is the link length, $q_i$ is the joint rotational angle, $\alpha_i$ is the twist angle, and $d_i$ is the offset between consecutive links. The values for these parameters are determined following the procedure outlined in [29].

To compute the transformation from the end-effector frame $O_6X_6Y_6Z_6$ (denoted as {E}) to the base frame $O_0X_0Y_0Z_0$ (denoted as {O}), we multiply the individual transformations along the kinematic chain:

$$T_6^0 = A_1^0 A_2^1 A_3^2 A_4^3 A_5^4 A_6^5 \tag{15}$$

Furthermore, the transformation matrix from the base frame {O} to the end-effector frame {E} can be derived by taking the inverse of $T_6^0$:

$$T_0^6 = (T_6^0)^{-1} \tag{16}$$

If a point $P^O$ is defined in the base frame, its coordinates in the end-effector frame $P^E$ can be found using:

$$P^E = T_0^6 P^O \tag{17}$$

Equation (17) describes a coordinates transformation process from the base frame to the end-effector frame.

### 4.3. Eye-to-Hand Transformation Model

Assuming the camera remains fixed relative to the robot's end-effector, we introduce a constant transformation matrix $T_E^C$ that maps points from the end-effector frame {E} to the camera frame {C}. The process of finding the matrix $T_E^C$ is called Eye-to-Hand Calibration.

In general, the goal is to solve for $T_E^C$, such that:

$$A_i^j \cdot T_E^C = T_E^C \cdot B_i^j \tag{18}$$

Here, $A_i^j$ is transformation of the end-effector between robot poses $i$ to pose $j$, and $B_i^j$ represents the transformation of a fixed object (e.g., a marker) as observed in the camera frame during the same motion.

These transformations are computed as:

$$A_i^j = (T_0^{6,i})^{-1} \cdot T_0^{6,j} \tag{19}$$

$$B_i^j = (T_M^{C,i})^{-1} \cdot T_M^{C,j} \tag{20}$$

Where $T_0^{6,i}$ and $T_0^{6,j}$ denote the poses of the end-effector at robot configurations $i$ and $j$, measured in the base frame, as defined in Equation (17). $T_M^{C,i}$ and $T_M^{C,j}$ denote the transformations of a marker M observed in the camera frame at the same respective robot poses. Further details are provided in Appendix C.

Various methods have been proposed to accurately estimate $T_E^C$ using multiple $(A_i^j, B_i^j)$ pairs. While calibration techniques are not the primary focus of this paper, several recent approaches are summarized for reference:

Kim et al. [30] presented a technique for eye-on-hand configurations, deriving a closed-form solution for equation (18) using two robot movements. Evangelista et al. [31] introduced a unified iterative method based on reprojection error minimization, applicable to both eye-on-base and eye-in-hand setups. This approach does not require explicit camera pose estimation and demonstrates improved accuracy compared to traditional methods. Strobl & Hirzinger [32] proposed an optimal calibration method for eye-in-hand systems, utilizing a novel metric on SE(3) and an error model for nonlinear optimization. This method adapts to system precision characteristics and performs well in equation (18) formulation.

### 4.4. Robot-to-Camera Kinematic Model

For a stereo camera system, this camera frame is located at the center of the stereo baseline, as shown in Figure 6. The coordinates of a point in space, measured in the base frame, can then be expressed in the camera frame as:

$$P^C = T_E^C T_0^6 \ P^O \tag{21}$$

Equation (21) describes how a given 3D point in the base frame $P^O = [X^O, Y^O, Z^O]$ is mapped to the camera frame $P^C = [X^C, Y^C, Z^C]$ using transformation:

$$\text{Robot manipulator mapping } \mathcal{H}: \ P^O = [X^O, Y^O, Z^O]^T \rightarrow P^C = [X^C, Y^C, Z^C]^T \tag{22}$$

The mapping function $\mathcal{H}$ is nonlinear and depends on the current joint angle of robot $q = [q_i | i \in 1,2,3,4,5,6]$, robot geometric parameter $P_{a\_robot} = [a_i, \ \alpha_i, \ d_i \ | i \in 1,2,3,4,5,6]$, and constant transformation matrix $T_E^C$.

### 4.5. Eye-in-hand Kinematic Model

By combining the stereo camera mapping $M$ from Equation (11) and the robot manipulator mapping $\mathcal{H}$ from Equation (22), we define a nonlinear transformation $\mathcal{F}$, which maps any point measured in the base frame {O} to its image coordinates as captured by the stereo camera. This transformation is expressed as:

$$\text{Eye-in-hand mapping } \mathcal{F}: \ P^O = [X^O, Y^O, Z^O]^T \rightarrow p_{image} = [u_l, u_r, v]^T \tag{23}$$

The Mapping $\mathcal{F}$ is nonlinear and depends on variables current joint angles: $q$ and parameters $P_a$, which includes stereo camera parameter $P_{a\_camera}$, robot geometric parameter $P_{a\_robot}$, and transformation matrix $T_E^C$. In other words:

$$\text{For any time } t \geq 0, \ p_{image} = \mathcal{F}(q(t), Pa, P^O) \tag{24}$$

$$Pa = [P_{a_{camera}}, P_{a_{robot}}, T_E^C] \tag{25}$$

### 4.6. Robot-to-Camera Inverse Kinematic Model

Inverse kinematics determines the joint angles required to achieve a given camera pose relative to the inertial frame. The camera pose in the inertial frame can be expressed as a 4X4 matrix:

$$Pose^O = \begin{bmatrix} n_x & s_x & a_x & d_x \\ n_y & s_y & a_y & d_y \\ n_z & s_z & a_z & d_z \\ 0 & 0 & 0 & 1 \end{bmatrix} \tag{26}$$

Here, the vectors $[n_x, n_y, n_z]^T$, $[s_x, s_y, s_z]^T$ and $[a_x, a_y, a_z]^T$ represent the camera's directional vectors for Yaw, Pitch, and Roll, respectively, in the base frame {C}. Additionally, the vector $[d_x, d_y, d_z]^T$ denotes the absolute position of the camera center in the base frame {C}.

The camera pose in the camera frame {C} is straightforward as it can be expressed as another 4X4 matrix:

$$Pose^C = \begin{bmatrix} 1 & 0 & 0 & 0 \\ 0 & 1 & 0 & 0 \\ 0 & 0 & 1 & 0 \\ 0 & 0 & 0 & 1 \end{bmatrix} \tag{27}$$

The nonlinear inverse kinematics problem involves solving for the joint angles $q$ that satisfy the equation:

$$Pose^C = T_E^C \cdot T_0^6(q) \cdot Pose^O \qquad (28)$$

where $T_E^C$ is the transformation from the end-effector frame to the camera frame, and $T_0^6(q)$ represents the transformation from the base frame to the end-effector frame, which is a function of the joint angles $q$.

The formulas for computing each joint angle are derived from the geometric parameters of the robot. The results of the inverse kinematics calculations for the ABB IRB 4600 elbow manipulator [33], used for simulations in this paper, are summarized in Appendix A.

### 4.7. Robot Dynamic Model

Dynamic models are included in the inner joint control loop, which will be discussed in section 6. Without derivation, the dynamic model of a serial of 6-link rigid, non-redundant, fully actuated robot manipulator can be written as [34]:

$$(D(q) + J)\ddot{q} + (C(q, \dot{q}) + \frac{B}{r})\dot{q} + g(q) = u \qquad (29)$$

Where $q \in \mathbb{R}^{6X1}$ is the vector of joint positions, and $u \in \mathbb{R}^{6X1}$ is the vector of electrical power input from DC motors inside joints, $D(q) \in \mathbb{R}^{6X6}$ is the symmetric positive defined matrix, $C(q, \dot{q}) \in \mathbb{R}^{6X6}$ is the vector of centripetal and Coriolis effects, $g(q) \in \mathbb{R}^{6X1}$ is the vector of gravitational torques, $J \in \mathbb{R}^{6X6}$ is a diagonal matrix expressing the sum of actuator and gear inertias, $B \in \mathbb{R}^{6X1}$ is the damping factor, $r \in \mathbb{R}^{6X1}$ is the gear ratio.

## 5. Control Policy Diagram

Figure 7 illustrates the overall control system architecture, designed to guide the robot manipulator so that the camera reaches its desired pose, $\overline{pose^O} \in \mathbb{R}^{4X4}$, in the world space. This architecture follows a similar structure to that used in prior work on tool trajectory control for robotic manipulators [35]. To facilitate visual servoing, three fiducial markers are placed within the camera's field of view, with their coordinates in the inertial frame pre-determined and represented as $P^O \in \mathbb{R}^{9X1}$. Using the robot's inverse kinematics and the eye-in-hand kinematic model, the expected image coordinates of these fiducial markers, when viewed from the desired camera pose, are computed as $\overline{p_{image}} \in \mathbb{R}^{9X1}$. These computed image coordinates serve as reference targets in the feedback control loop.

The IBVS framework is implemented within a cascaded feedback loop. In the outer control loop, the camera controller processes the visual feedback error, $e_p$, which represents the difference between the image coordinates of the fiducial points at the current and desired camera poses. Based on this error, the outer loop generates reference joint angles, $q_{feedback} \in \mathbb{R}^{6X1}$, to correct the robot's configuration.

The inner control loop, shown in Figure 8, incorporates the robot's dynamic model to regulate the joint angles, ensuring they align with the commanded reference angles, $q_{ref} \in \mathbb{R}^{6X1}$. However, due to the limitations of low-fidelity and inexpensive joint encoders, as well as inherent dynamic errors such as joint compliance, high frequency noises and low frequency model disturbances are introduced into the system. All sources of errors from the joint control loop are collectively modeled as an input disturbance, $d_{q_T} \in \mathbb{R}^{6X1}$, which affects the outer control loop.

The feedforward control loop operates as an open-loop system, quickly bringing the camera as close as possible to its target pose, despite the presence of input disturbances.

The feedforward controller outputs a reference joint angle command, $q_{feedforward} \in \mathbb{R}^{6X1}$, which is sent to the inner loop to facilitate rapid convergence to the desired configuration.

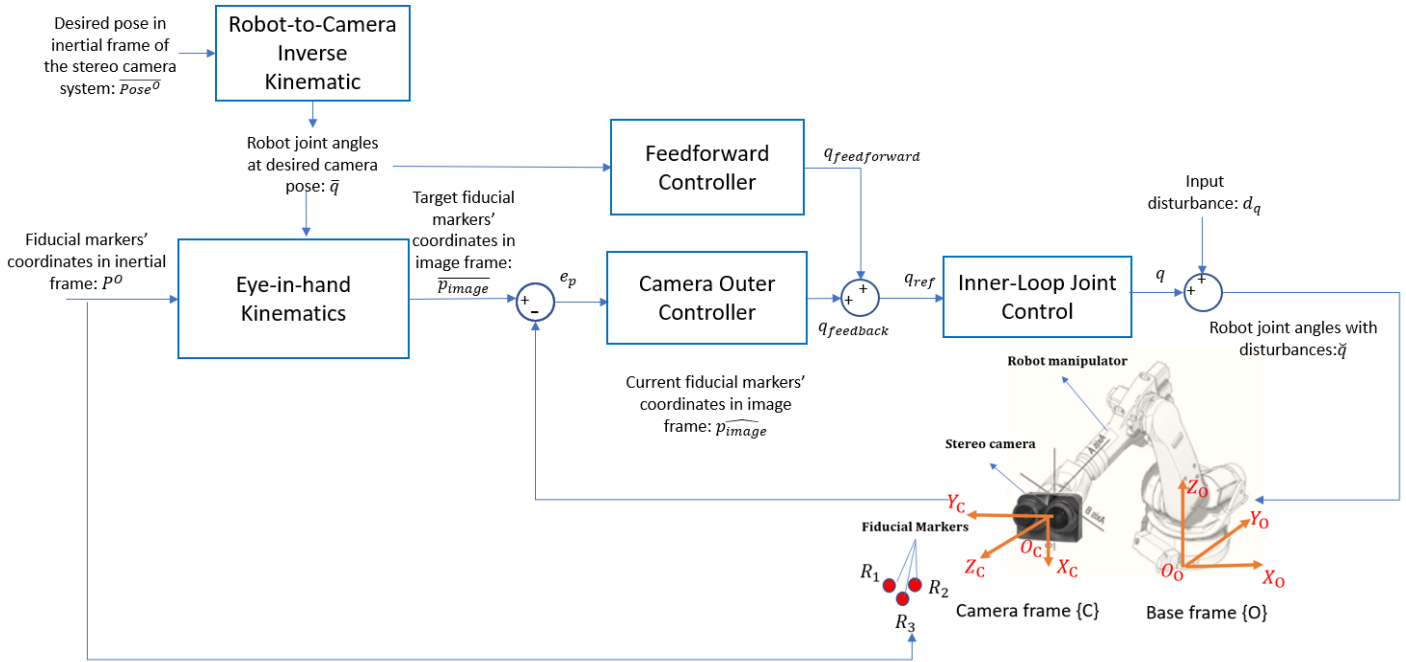

**Figure 7. Feedforward-feedback control architecture.** *Note: In mathematics, the in-loop hardware is equivalent to the Eye-in-hand Kinematics Model.*

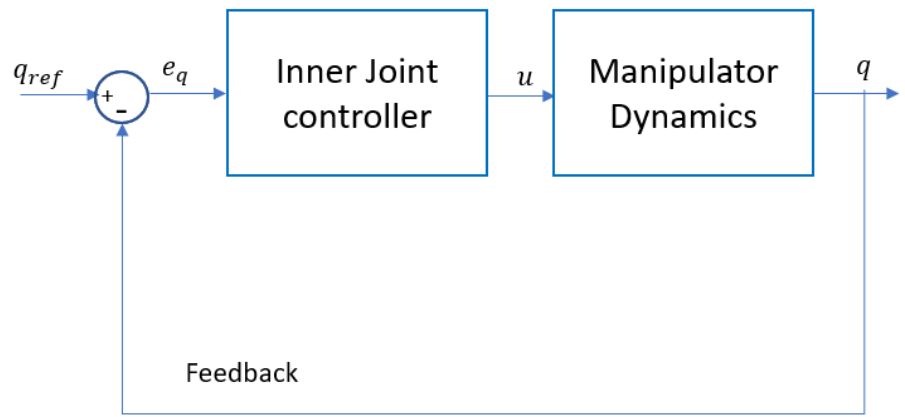

**Figure 8.** Inner joint angle control loop

## 6. Controller Designs

### 6.1. Inner Joint Angle Control Loop

As illustrated in Figure 8, the primary objective of the inner joint controller is to stabilize the manipulator dynamics, represented by the nonlinear equation (29). This dynamic model can be simplified to the following form:

$$M(q)\ddot{q} + h(\mathrm{q}, \dot{\mathrm{q}}) = \mathrm{u} \qquad (30)$$

Where:

$$M(q) = D(q) + J \tag{31}$$

$$h(\mathrm{q}, \dot{\mathrm{q}}) = (C(q, \dot{q}) + \frac{B}{r})\dot{q} + \mathrm{g}(q) \tag{32}$$

To facilitate control design, the input $u$ is redefined as::

$$u = M(q)v + h(\mathrm{q}, \dot{\mathrm{q}}) \tag{33}$$

Here $v \; \epsilon \mathbb{R}^{6X1}$ is a virtual input. Substituting Equation (33) into Equation (30), and noting that $M(q)\epsilon \, \mathbb{R}^{6X6}$ is invertible, the system dynamics simplify to:

$$\ddot{q} = v \tag{34}$$

This transformation technique is known as feedback linearization, resulting in a system of six decoupled double integrators, as described in Equation (34). The entire feedback linearization framework is illustrated in Figure 9. In this control architecture, the actual joint angles $q \; \epsilon \mathbb{R}^{6X1}$ are regulated to follow the desired joint angles $q_R \; \epsilon \mathbb{R}^{6X1}$.

A nonlinear interface block computes the actual control input $u$ $\epsilon \; \mathbb{R}^{6X1}$ from the virtual control input $v \; \epsilon \mathbb{R}^{6X1}$ using Equation (33). This block also calculates the matrices $M(q) \; \epsilon \; \mathbb{R}^{6X6}$, and $h(\mathrm{q}, \dot{\mathrm{q}}) \; \epsilon \; \mathbb{R}^{6X1}$ based on the current joint angles $q$ and their velocities $\dot{q}$. The linear joint controller is designed using Youla parameterization technique [27] to control the nominally linear system in Equation (34).

Since the six equations in (34) are decoupled and identical in structure, a Single-Input Single-Output (SISO) controller can be first designed and then extended to a Multiple-Input Multiple-Output (MIMO) controller by duplication. The SISO version is defined as:

$$v_{SISO} = \ddot{q}_{SISO} \tag{35}$$

Let the SISO controller be denoted $G_{C_{MIMO}}^{Inner}$, then the MIMO controller is:

$$G_{C_{MIMO}}^{Inner} = G_{C_{SISO}}^{Inner} \cdot I_{6X6} \tag{36}$$

where $I_{6X6}$ is a $6 \times 6$ identity matrix. The transfer function of the SISO plant, derived from Equation (34), is:

$$G_{P_{SISO}}^{Inner} = \frac{1}{s^2} \tag{37}$$

This system has two Bounded Input Bounded Output (BIBO) unstable poles at the origin. To ensure internal stability of the closed-loop system, the transfer function $T_{SISO}$, must satisfy the interpolation conditions [36]:

$$T_{SISO}^{inner} \; (s = 0) = 1 \tag{38}$$

$$\frac{dT_{SISO}^{inner}}{ds}|_{s=0} = 0 \tag{39}$$

Using the relationship:

$$T_{SISO}^{inner} = Y_{SISO}^{inner} G_{P_{SISO}}^{Inner} \tag{40}$$

the Youla parameter $Y_{SISO}^{inner}$ can be derived from a properly chosen $T_{SISO}^{inner}$ that satisfies Equations (38) and (39). Without delving into the full design details, a candidate closed-loop transfer function that meets these conditions is:

$$T_{SISO}^{inner} = \frac{(3\tau_{in}s + 1)}{(\tau_{in}s + 1)^3} \tag{41}$$

Where $\tau_{in}$ determines the location of poles and zeros, and thus the system's bandwidth. Tuning $\tau_{in}$ allows the controller to achieve a desired tradeoff between speed and overshoot.

The next step is to derive $G_{C_{SISO}}^{Inner}$ from relationships between the closed-loop transfer function, $T_{SISO}^{inner}$, the sensitivity transfer function, $S_{SISO}^{inner}$, and the Youla transfer function, $Y_{SISO}^{inner}$, in Equations (42)-(44):

$$Y_{SISO}^{inner} = T_{SISO}^{inner} G_{p_{SISO}}^{Inner^{-1}} = \frac{s^2(3\tau_{in}^2s + 1)}{(\tau_{in}s + 1)^3} \tag{42}$$

$$S_{SISO}^{inner} = 1 - T_{SISO}^{inner} = \frac{s^2(\tau_{in}^3s + 3\tau_{in}^2)}{(\tau_{in}s + 1)^3} \tag{43}$$

$$G_{C_{SISO}}^{Inner} = Y_{SISO}^{inner} S_{SISO}^{inner^{-1}} = \frac{3\tau_{in}^2s + 1}{\tau_{in}^3s + 3\tau_{in}^2} \tag{44}$$

Substituting back into Equation (36), the resulting MIMO controller is:

$$G_{C_{MIMO}}^{Inner} = \frac{3\tau_{in}^2s + 1}{\tau_{in}^3s + 3\tau_{in}^2} \cdot I_{6\times6} \tag{45}$$

The final expression for the closed-loop MIMO system is:

$$T_{MIMO}^{inner} = \frac{(3\tau_{in}s + 1)}{(\tau_{in}s + 1)^3} \cdot I_{6\times6} \tag{46}$$

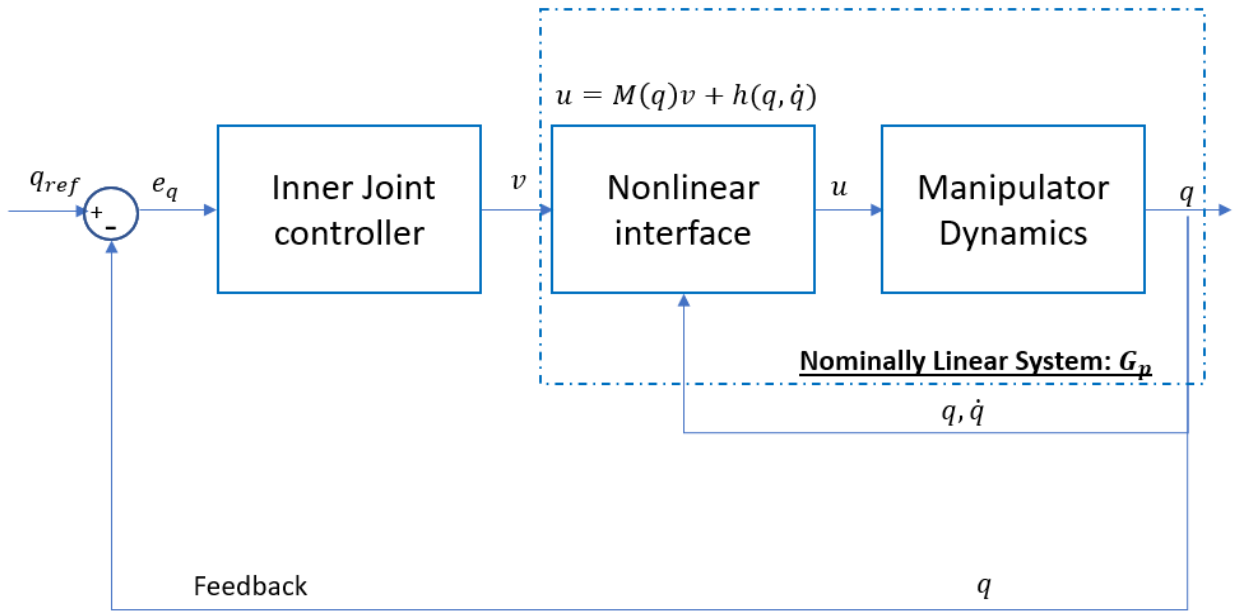

**Figure 9.** Feedback linearization Youla control design for inner loop

## 6.2. Feedforward Control Loop

The feedforward control loop operates in open-loop form and is subject to input disturbances, as depicted in Figure 10. The feedforward controller is designed to approximate the inverse dynamics of the inner-loop joint controller, whose closed-loop transfer function is given in Equation (46). Therefore, the feedforward transfer function is constructed as the inverse of the closed-loop system:

$$T_{forward} = \frac{1}{T_{inner-closed}} \ \frac{1}{(\tau_{forward}s+1)^2} = \frac{(\tau_{in}s+1)^3}{(3\tau_{in}s+1)} \ \frac{1}{(\tau_{forward}s+1)^2} \cdot I_{6\times6} \qquad (47)$$

Substituting the closed-loop response from Equation (46), we obtain:

$$T_{forward} = \frac{(\tau_{in}s+1)^3}{(3\tau_{in}s+1)} \ \frac{1}{(\tau_{forward}s+1)^2} \cdot I_{6\times6} \qquad (48)$$

Here, the additional double poles at s = -1/$\tau_{forward}$ are introduced to ensure that the feedforward transfer function is proper (i.e., physically realizable). To minimize the impact on bandwidth while ensuring realizability, the double poles are placed significantly faster than the original system dynamics. Specifically, the time constant $\tau_{forward}$ is chosen to be one-tenth of the inner-loop time constant:

$$\tau_{forward} = 0.1\tau_{in} \qquad (49)$$

This design ensures that the feedforward action closely approximates the inverse of the inner-loop dynamics without introducing high-frequency noise amplification.

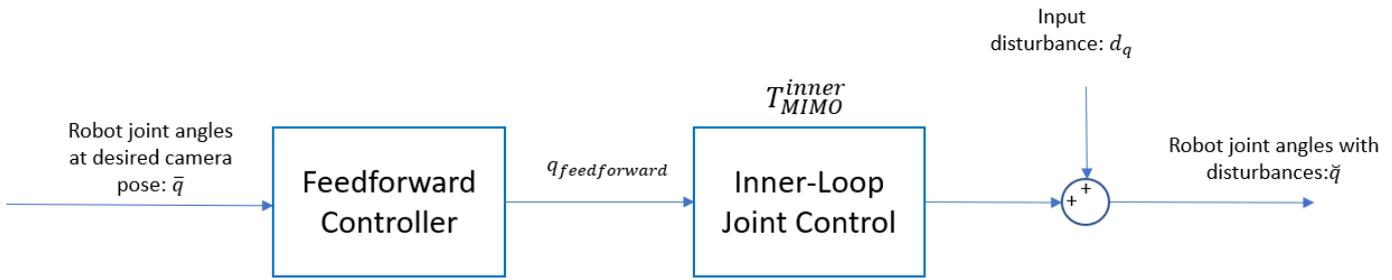

**Figure 10.** Feedforward control design

### 6.3. Outer Feedback Control Loop

In the control system block diagram (Figure 7), the closed-loop linear transfer function is derived in Equation (46), while the eye-in-hand kinematic model is defined as a nonlinear mapping in Equations (24) and (25).

When the input to this model is a set of perturbed joint angles $\check{q}(t)$, the output corresponds to the estimated image coordinates $\widehat{p_{image}}$. Revising Equation (24) accordingly, we express the model as:

$$\widehat{p_{image}} = \mathcal{F}(\check{q}(t), Pa, P^O) \tag{50}$$

Here, $P^O \epsilon \, \mathbb{R}^{9X1}$, denotes the coordinates of fiducial markers in the inertial frame, and $Pa$ includes fixed parameters such as camera intrinsics, robot kinematics, and the transformation matrix from the end-effector to the camera frame.

To linearize this nonlinear model, we select a nominal joint configuration $\check{q}^0 \epsilon \, \mathbb{R}^{6X1}$, and approximate the function $\mathcal{F}$ using a first-order Taylor expansion, resulting in the following linearized form:

$$\widehat{p_{image}} \approx J(\check{q}^0, Pa, P^O)\check{q}(t) + \mathcal{F}(\check{q}^0, Pa, P^O) \tag{51}$$

Where $J(\check{q}^0, Pa, P^O) \, \epsilon \, \mathbb{R}^{6X6}$ is the Jacobian matrix of $\mathcal{F}$ evaluated at $\check{q}^0$. Letting $C_1 = J(\check{q}^0, Pa, P^O)$, $C_2 = \mathcal{F}(\check{q}(t), Pa, P^O)$, Equation (51) can be rewritten as:

$$\widehat{p_{image}} = C_1\check{q}(t) + C_2 \tag{52}$$

Defining the offset-corrected output as $\widehat{p_{image}}' = \widehat{p_{image}} - C_2$, we obtain a purely linear relation that simplifies analysis. The complete linearized feedback loop is illustrated in Figure 11.

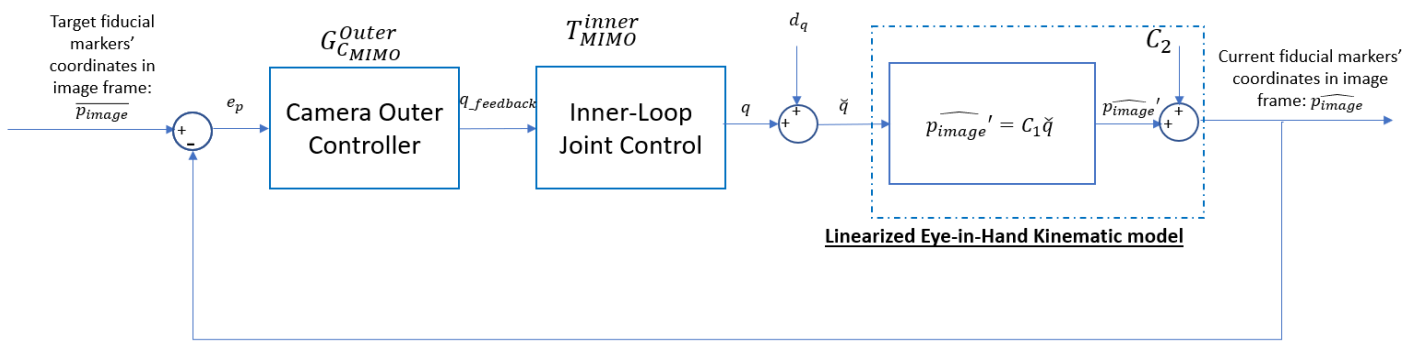

**Figure 11.** Feedback loop with linearized model

The resulting linear plant transfer function is given by:

$$G_{p_{MIMO\_outer}}^{linear} = \frac{\widehat{p_{image}}'}{q_{feedback}} = C_1 \frac{(3\tau_{in}s + 1)}{(\tau_{in}s + 1)^3} \cdot I_{6\times6} \tag{53}$$

Since $C_1$ introduces coupling, he first step toward designing an observer for this multivariable system is to derive its Smith-McMillan form [36]. This can be achieved via Singular Value Decomposition (SVD):

$$G_{p_{MIMO\_outer}}^{linear} = U_L M_p U_R \tag{54}$$

where $U_L \epsilon \mathbb{R}^{9X9}$ and $U_R \epsilon \mathbb{R}^{6X6}$ are the left and right unimodular matrices, and $M_p \epsilon \mathbb{R}^{9X6}$ is the Smith-McMillan form a diagonal matrix of transfer functions. Each nonzero diagonal element takes the form:

$$M_p(i,i) = gain(i) \cdot \frac{(3\tau_{in}s+1)}{(\tau_{in}s+1)^3}, i\epsilon(1,2,3,4,5,6) \tag{55}$$

Where $gain \; \epsilon \; \mathbb{R}^{6X1}$ is a numerical vector.

The Youla controller design for each decoupled entry in $M_p$ is straightforward, as the system is stable (i.e., all poles and zeros lie in the left-half plane). The Youla parameterization allows shaping of the closed-loop dynamics via a new transfer function $M_Y$, which modifies the system's poles and zeros. A second-order Butterworth filter is selected for the desired closed-loop dynamics:

$$M_T = \frac{\omega_n^2}{(s^2 + 2\zeta\omega_n s + \omega_n^2)} \cdot \begin{bmatrix} I_{6\times6} & Zero_{3\times3} \\ Zero_{3\times3} & Zero_{3\times3} \end{bmatrix}, \tag{56}$$

Here $\omega_n$ is the natural frequency that determines the bandwidth of the outer loop (ensuring $1/\omega_n > \tau_{in}$ to maintain proper hierarchy with the inner loop), and $\zeta$ is the damping ratio for tuning response smoothness. The last 3 coordinates remain uncontrolled in the feedback loop.

The corresponding decoupled Youla controller is:

$$M_Y(i,i) = \frac{M_T(i,i)}{M_p(i,i)} = \frac{1}{gain(i)} \frac{\omega_n^2}{(s^2+2\zeta\omega_n s+\omega_n^2)} \frac{(\tau_{in}s+1)^3}{(3\tau_{in}s+1)}, , i\epsilon(1,2,3,4,5,6) \tag{57}$$

Following Equations (43) –(45), the final coupled Youla, closed loop, sensitivity, and observer transfer function matrices are computed as:

$$Y_{MIMO\_outer}^{linear} \epsilon \; \mathbb{R}^{6X9} = U_R M_Y U_L \tag{58}$$

$$T_{y_{MIMO\_outer}}^{linear} \epsilon \; \mathbb{R}^{9X9} = G_{p_{MIMO\_outer}}^{linear} \cdot Y_{MIMO\_outer}^{linear} \tag{59}$$

$$S_{y_{MIMO\_outer}}^{linear} \epsilon \; \mathbb{R}^{9X9} = 1 - T_{y_{MIMO\_outer}}^{linear} \tag{60}$$

$$G_{C_{MIMO\_outer}}^{linear} \epsilon \; \mathbb{R}^{6X9} = Y_{MIMO\_outer}^{linear} \cdot (S_{y_{MIMO\_outer}}^{linear})^{-1} \tag{61}$$

This controller is based on a linearization about a fixed point $\breve{q}^0$, and is therefore valid only within a neighborhood of that configuration. As the actual joint angles $\breve{q}(t)$ deviate from $\check{q}^0$, the mismatch between the linearized model (Equation 51) and the true nonlinear model (Equation 50) grows.

To address this, an adaptive control strategy is proposed, illustrated in Figure 12. This method updates the controller in real-time using the current estimated joint configuration. First, joint angles $\tilde{q}$ are estimated from the current image coordinates:

$$\tilde{q} = \mathcal{F}^{-1}(\widehat{p_{image}}, Pa, P^O) \tag{62}$$

Here，$\mathcal{F}^{-1}$ is the inverse of the nonlinear eye-in-hand model, which includes both coordinate transformation from the image to the end-effector frame and the inverse kinematics of the robot.

Using $\tilde{q}$, the current Jacobian and updated SVD-based decomposition are recalculated each time step. The Smith-McMillan form and corresponding controller $G_{C_{MIMO\_outer}}^{linear}$ are then reconstructed via Equations (58)–(61), yielding an adaptive Youla controller that remains effective across varying joint configurations.

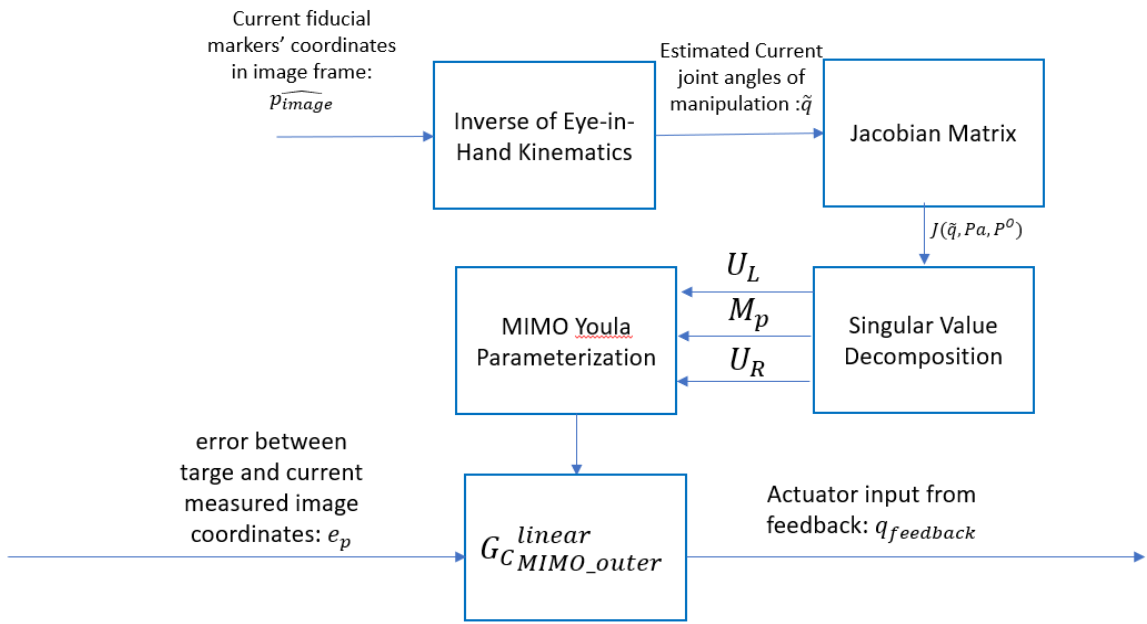

**Figure 12.** Adaptive feedback loop

## 7. Simulations Results

To evaluate the performance of our controller design, we simulated two scenarios in MATLAB Simulink using a Zed 2 stereo camera system [37] and an ABB IRB 4600 elbow robotic manipulator [33]. All simulations are conducted entirely in Simulink, with hardware-in-the-loop components replaced by their corresponding mathematical models. The specifications for the camera system and robot manipulator are summarized in the tables in the appendix. The camera system performs 2D feature estimation of three virtual points in space, with their coordinates in the inertial frame selected as: $[-0.5m\ 0\ 0]^T$, $[0\ 0\ 0.5m]^T$, $[2m, -2m, 0]^T$.

Many camera noise removal algorithms have been proposed and shown to be effective in practical applications, such as spatial filters [38], wavelet filters [39], and the image averaging technique [40]. Among these denoising methods, there is always a tradeoff between computational efficiency and performance. For this paper, we assume that the images captured by the camera have been preprocessed using one of these methods, and the noise has been almost perfectly attenuated. In other words, the only remaining disturbances in the system are due to unmodeled joint dynamics, such as compliance and flexibility, which are modeled as input disturbances in the controlled system.

Two scenarios were simulated:

- Scenario 1: Without input disturbances.
- Scenario 2: With a 1° step input disturbance added to each joint of the robot arms for the entire simulation time.

In both scenarios, the camera system starts from an initial pose in the inertial frame, denoted as $Pose^O_{initila}$, and maneuvers target pose, denoted as $Pose^O_{final}$. Table 1 summarizes the initial and final poses for each scenario, along with the corresponding joint configurations.

Figure 13 and Figure 15 present the responses of the six joint angles for the two scenarios, respectively. Figure 14 and Figure 16 show the responses of the nine image coordinates over time for each scenario. In each case, the feedback-only controlled system (left plot) is compared to the feedforward-and-feedback controlled system (right plot). These comparisons focus on overshoot, response time, and target tracking performance.

**Table 1.** Camera Pose and Robot Joint Angles at Initial and Final State of Simulation Scenarios

| Format | Camera Pose | | Robot Joint Angles | |
|---|---|---|---|---|
| | $\begin{bmatrix} n_x & s_x & a_x & d_x \\ n_y & s_y & a_y & d_y \\ n_z & s_z & a_z & d_z \end{bmatrix}$ <br><br> Where $[n_x, n_y, n_z]^T$, $[s_x, s_y, s_z]^T$ and $[a_x, a_y, a_z]^T$ are Yaw, Pitch, and Roll, and $[d_x, d_y, d_z]^T$ (in meters) is the position, measured in inertial frame {O}. | | $\begin{bmatrix} q_1 & q_2 \\ q_3 & q_4 \\ q_5 & q_6 \end{bmatrix}$ <br><br> Where $[q_1, q_2, q_3, q_4, q_5, q_6]$ in (degrees) are robot joint angles. | |
| | ***Scenario* 1** | | ***Scenario* 2** | |
| | Camera Pose | Robot Joint Angles | Camera Pose | Robot Joint Angles |
| Initial State | $\begin{bmatrix} 0 & 0 & 1 & 1.27 \\ 0 & 1 & 0 & 0 \\ -1 & 0 & 0 & 1.57 \end{bmatrix}$ | $\begin{bmatrix} 0° & 0° \\ 0° & 0° \\ 0° & 0° \end{bmatrix}$ | $\begin{bmatrix} -0.070 & -0.998 & 0.002 & 1.064 \\ -0.996 & 0.070 & -0.060 & 0.964 \\ -0.060 & -0.007 & -0.998 & 0.939 \\ 0 & 0 & 0 & 1 \end{bmatrix}$ | $\begin{bmatrix} 42.40° & 21.20° \\ 4.58° & -2.86° \\ 66.46° & -42.40° \end{bmatrix}$ |
| Final State | $\begin{bmatrix} -0.11 & 0.14 & 0.98 & 1.31 \\ -0.09 & 0.98 & -0.15 & -0.01 \\ -0.99 & -0.10 & -0.09 & 1.49 \end{bmatrix}$ | $\begin{bmatrix} 0.48° & 2.21° \\ 2.05° & -82.68° \\ 9.46° & 77.47° \end{bmatrix}$ | $\begin{bmatrix} 0 & -1 & 0 & 1 \\ -1 & 0 & -1 & 1 \\ 0 & 0 & 0 & 1 \end{bmatrix}$ | $\begin{bmatrix} 45° & 18.59° \\ 4.35° & 0° \\ 67.06° & -45° \end{bmatrix}$ |

The bandwidth of actuators in industrial robot manipulators typically ranges from 5 Hz to 100 Hz, corresponding approximately to 31 rad/s to 628 rad/s. For the purposes of this simulation, we assume that the inner loop—representing joint-level control—operates at a fixed bandwidth of 100 rad/s across all scenarios. This corresponds to a time constant of:

$$\tau_{in} = 0.01\text{s} \tag{63}$$

In cascaded control architectures, it is standard practice to design the inner loop to be at least 10 times faster than the outer loop to ensure proper time-scale separation and prevent interference. Accordingly, the outer-loop bandwidth is selected as one-tenth of the inner-loop bandwidth:

$$\omega_n = 10 \text{ rad/s} \tag{64}$$

Additional parameters for the controller are either computed analytically or selected based on design guidelines. In the feedforward controller, the time constant $\tau_{forward}$ is chosen to be one-tenth of the inner-loop time constant, as described in Equation (49):

$$\tau_{forward} = 0.1\tau_{in} = 0.001s \tag{65}$$

To ensure the outer-loop response is overdamped, and thus suppresses overshoot, the damping ratio $\zeta$ is selected to be relatively large:

$$\zeta = 10 \tag{66}$$

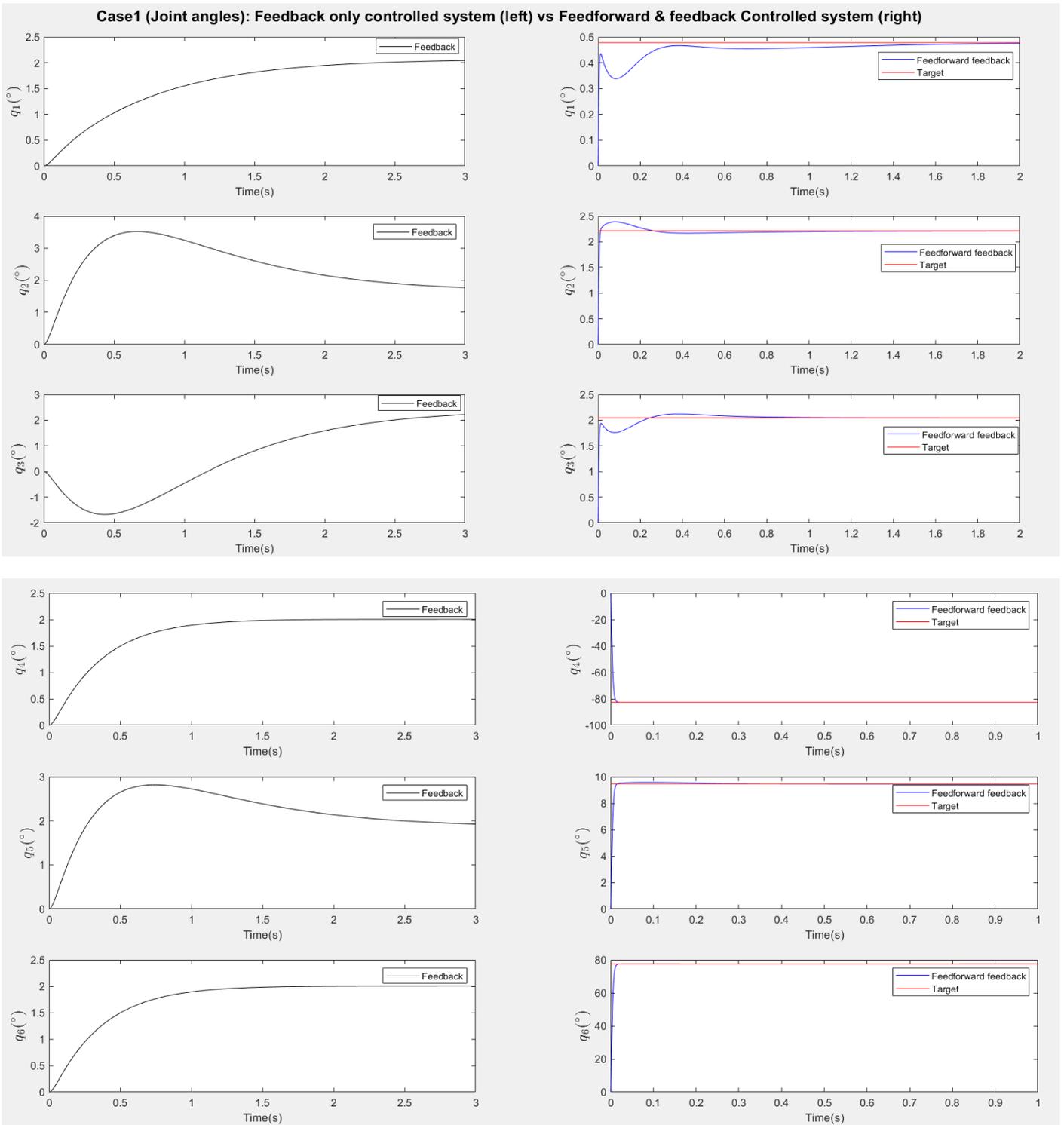

**Figure 13.** Scenario one (no disturbances): Response of robot joint angles. (*Left: Feed-back only responses, Right: Feedforward-feedback responses*).

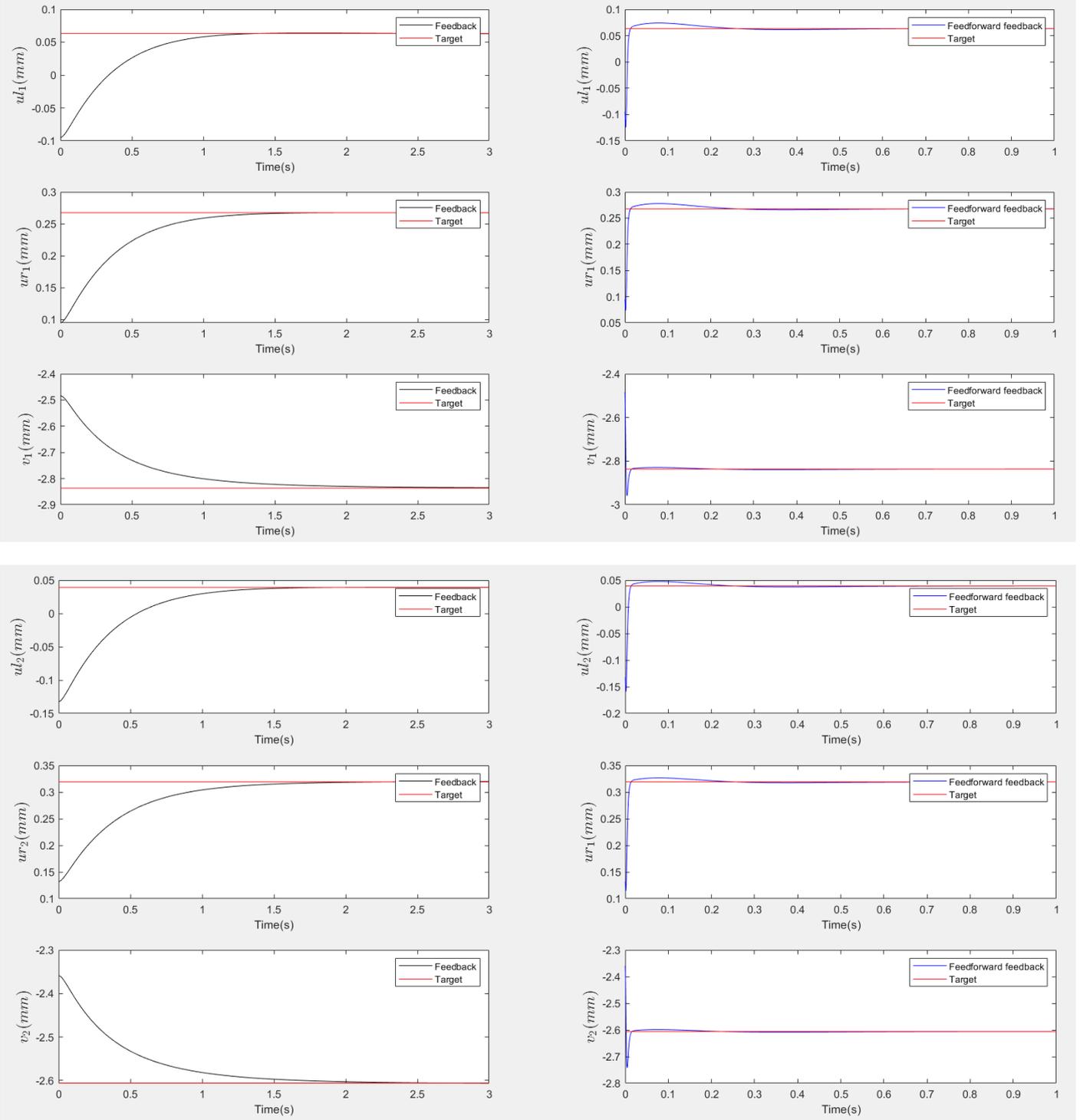

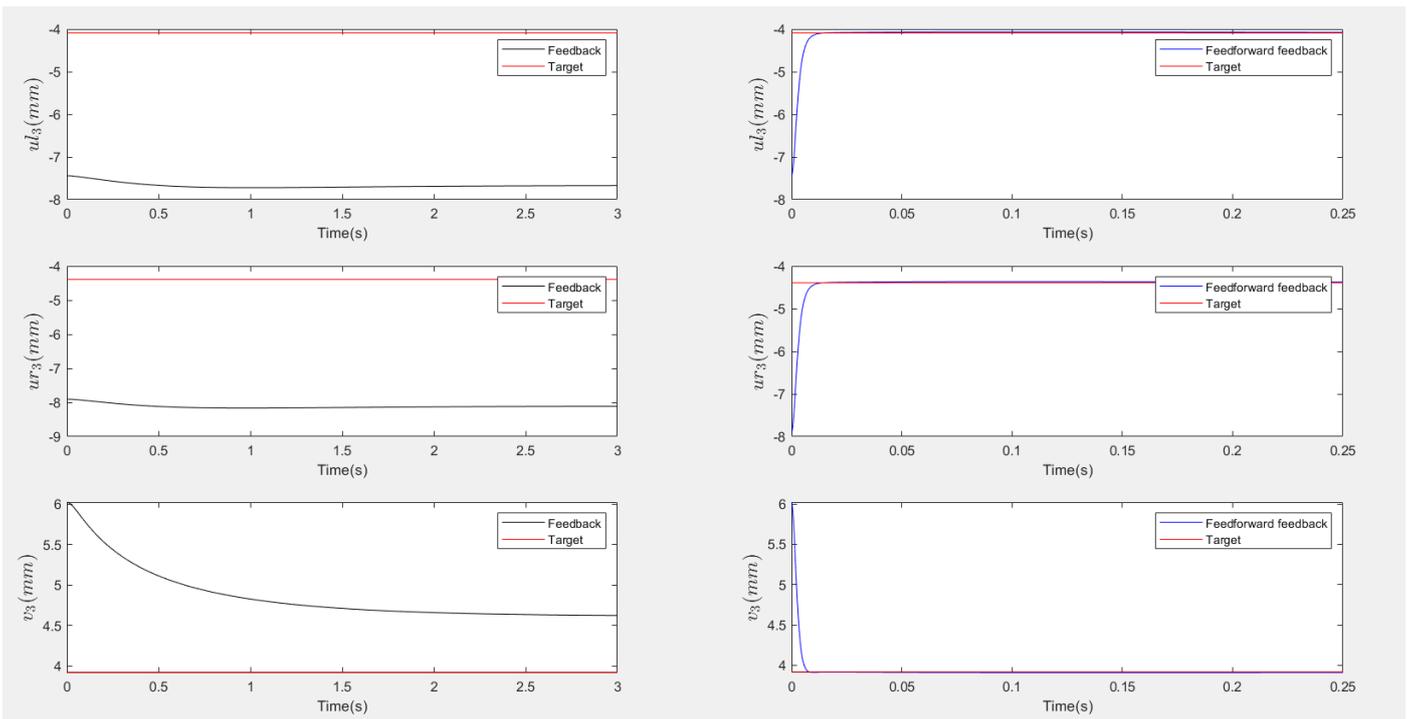

**Figure 14.** Scenario one (no disturbances): Response of image coordinates. (*Left: Feed-back only responses, Right: Feedforward-feedback responses).* The three coordinates of the third point are only matched in the feedforward-feedback approach.

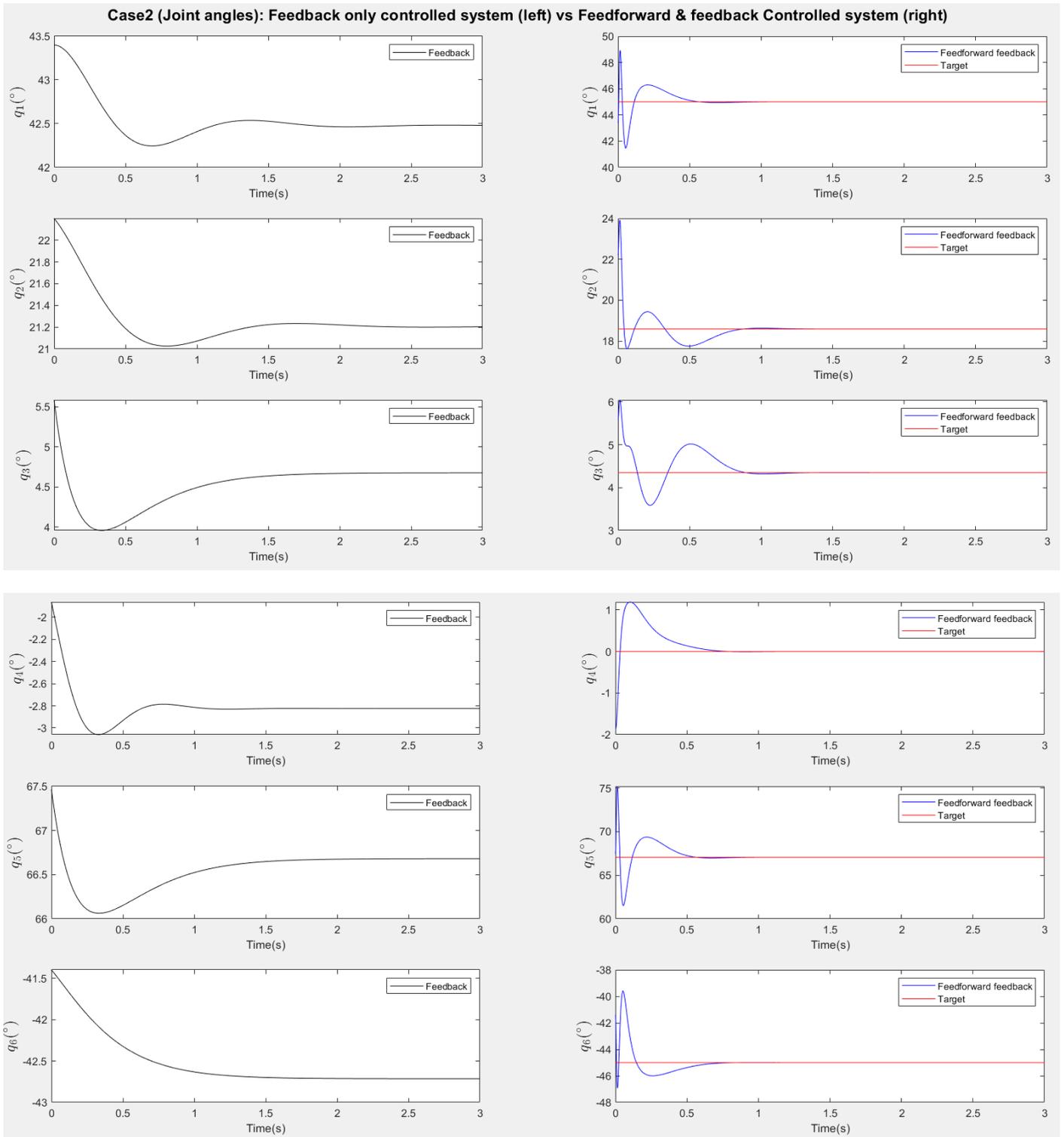

**Figure 15.** Scenario two (add disturbances): Response of robot joint angles. (*Left: Feed-back only responses, Right: Feedforward-feedback responses*).

**Case2 (Image Coordinates): Feedback only controlled system (left) vs Feedforward & feedback Controlled system (right)**

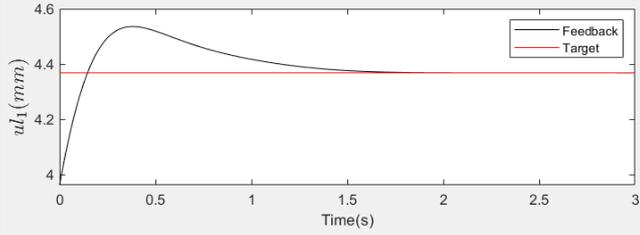 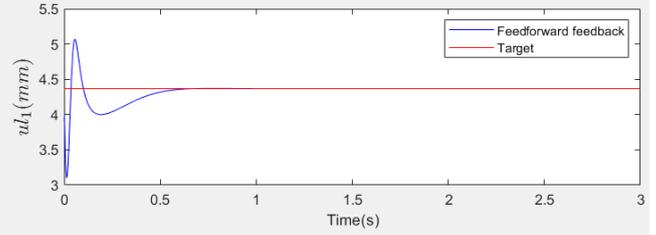

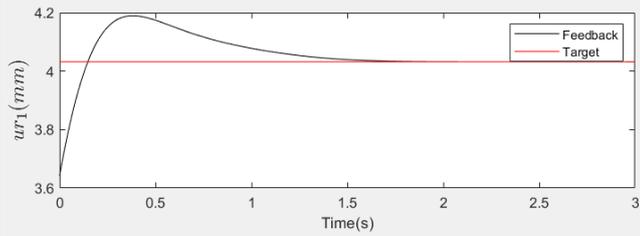 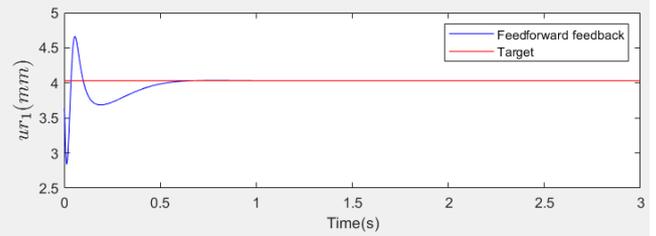

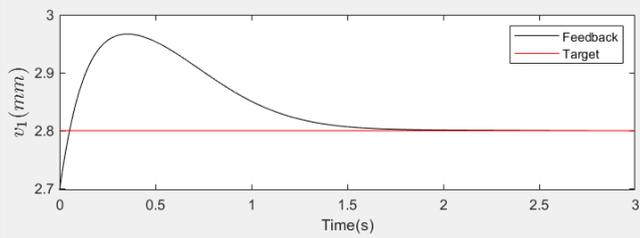 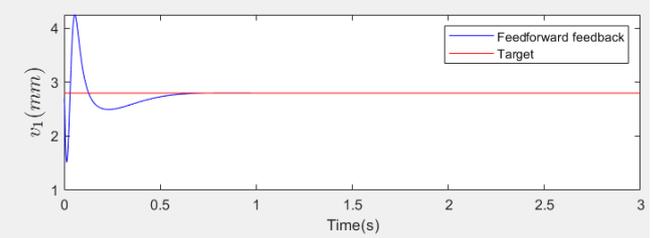

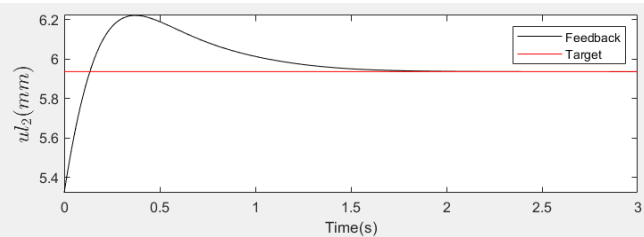 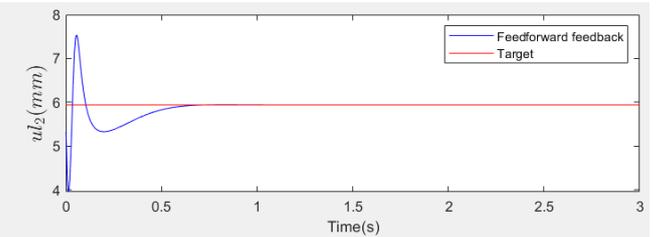

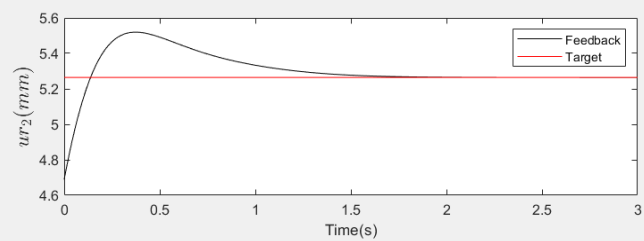 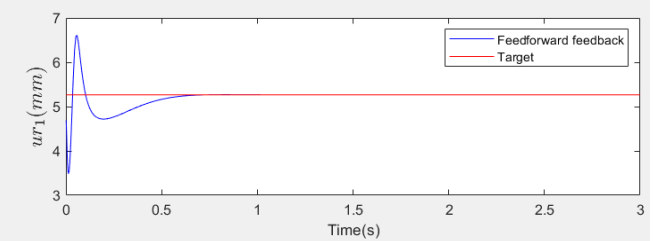

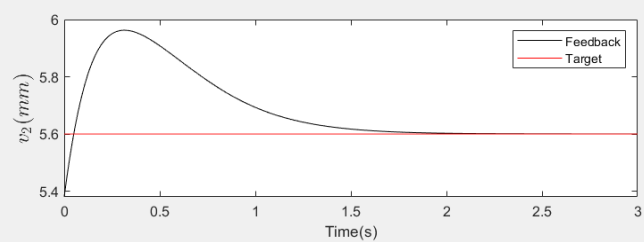 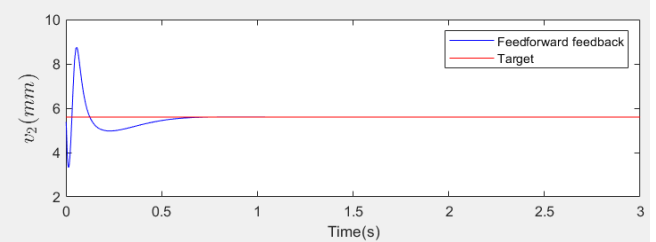

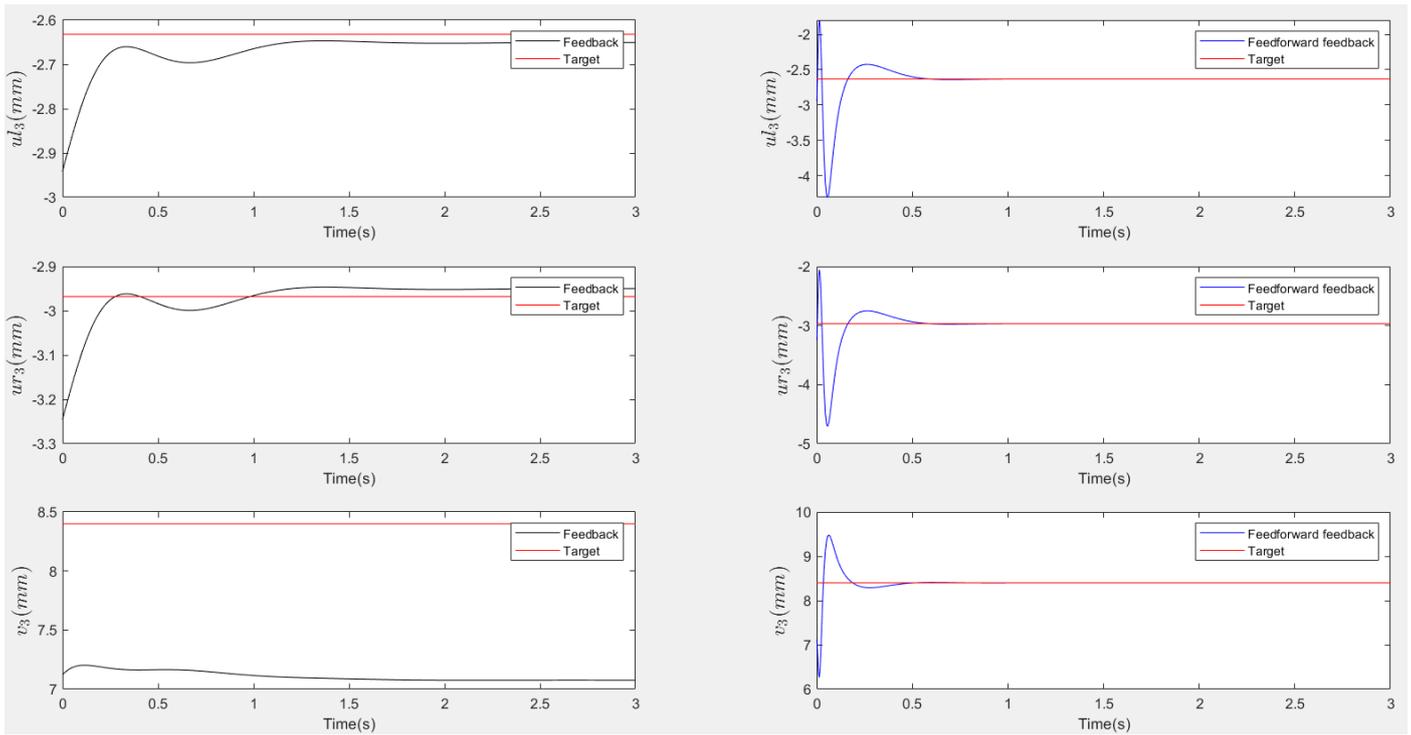

**Figure 16.** Scenario two (add disturbances): Response of image coordinates. (*Left: Feed-back only responses, Right: Feedforward-feedback responses*). The three coordinates of the third point are only matched in the feedforward-feedback approach.

The response plots indicate that both the feedback-only controller and the combined feedforward-and-feedback controller successfully stabilize the system and reach a steady state within three seconds, even in the presence of small input disturbances (Scenario Two). However, the feedback-only controller fails to guide the camera to its desired pose and falls into local minima, as evident from the third point's coordinates ($ul_{3}$, $ur_{3}$, $v_{3}$), which do not match the target at steady state. This issue arises from the overdetermined nature of the stereo-based visual servoing system, where the number of output constraints (9) exceeds the degrees of freedom (DoFs) available for control (6). As a result, the feedback controller can only match six out of nine image coordinates, leaving the rest coordinates unmatched.

In contrast, the feedforward-and-feedback controller avoids local minima and accurately moves the camera to the target pose. This is because the feedforward component directly controls the robot's joint angles rather than image features. Since the joint angles (6 DoFs) uniquely correspond to the camera's pose (6 DoFs), the feedforward controller helps the system reach the global minimum by using the desired joint configurations as inputs.

When comparing performance, the system with the feedforward controller exhibits a shorter transient period (less than 2 seconds) compared to the feedback-only system (less than 3 seconds). However, the feedforward controller can introduce overshoot, particularly in the presence of disturbances. This occurs because feedforward control provides an immediate control action based on desired setpoints, resulting in significant initial actuator input that causes overshoot. Additionally, a feedforward-only system is less robust against disturbances and model uncertainties. Fine-tuning the camera's movement under these conditions requires a feedback controller.

Therefore, the combination of feedforward and feedback control ensures fast and accurate camera positioning. The feedforward controller enables rapid convergence toward the desired pose, while the feedback controller improves robustness and corrects errors due to disturbances or uncertainties. Together, they work cooperatively to achieve optimal performance.

## 8. Conclusions

This work addresses a fundamental challenge in stereo-based image-based visual servoing (IBVS): the presence of local minima arising from the overdetermined nature of aligning 6-DOF camera poses with a higher number of image feature constraints. Through a formal reformulation of the stereo camera pose estimation problem as a Perspective-3-Point (P3P) case, we clarify the minimum information needed to uniquely resolve camera pose and highlight the risk of instability when controlling more image features than available degrees of freedom. To overcome this, a cascaded control framework was introduced, combining a feedforward joint-space controller with an adaptive Youla-parameterized feedback loop. The feedforward loop accelerates convergence by directly commanding joint configurations associated with the desired pose, while the feedback loop ensures robustness and fine-tuned tracking under disturbances or modeling uncertainties.

When compared to existing solutions, the proposed method offers a number of compelling advantages. Unlike hybrid switched controllers, which require explicit mode-switching logic and may suffer from discontinuities, our approach provides a unified framework that avoids local minima without controller mode changes. In contrast to 2½ D visual servoing, which blends pose and image features into a hybrid Jacobian, our design maintains the pure IBVS structure while preserving real-time feasibility. Moreover, compared to model predictive control (MPC) approaches that require high-frequency optimization, our strategy achieves comparable performance using linear control theory and avoids excessive computational overhead.

The table below summarizes key differences among four approaches:

**Table 2.** Comparative analysis of four visual servoing strategies addressing local minima and control robustness.

| Aspect | Hybrid Switched-System (Gans & Hutchinson, 2007) | 2½ D Visual Servoing (Chaumette & Malis, 2000) | MPC-Based IBVS (Roque et al., 2020) | Proposed Feedforward-feedback IBVS |
|---|---|---|---|---|
| Main Idea | Switch between IBVS and PBVS | Combine 2D image + 3D pose feature | Predict joint input from image error via MPC | Combine feedforward joint control with adaptive feedback |
| Local Minima Handling | Switching avoids image traps | Blended Jacobian avoids singularities | Predictive avoidance via optimization | Feedforward brings pose close to global minimum |
| Stability Guarantee | Local Lyapunov stability | Local/global under ideal pose estimation | Guaranteed via MPC feasibility | Inner/outer loop stability via Youla design |
| Computational Load | Moderate (switch logic) | Moderate (hybrid Jacobian computation) | High (real-time optimization) | Moderate (adaptive SVD and closed-form update) |
| Overshoot Risk | Low, but may switch abruptly | Minimal, smooth trajectories | Well-regulated by MPC | May occur in feedforward if not well tuned |
| Real-Time Suitability | Yes | Yes | Yes, With efficient solver | Yes |

Despite these advantages, several practical considerations must be acknowledged. The open-loop nature of the feedforward controller can introduce overshoot, particularly in systems with low damping or compliance in joints. While this is mitigated by tuning time constants and damping ratios, it underscores the importance of dynamic modeling accuracy. Additionally, the feedback controller's reliance on linearized approximations restricts global validity, though the proposed adaptive update strategy helps maintain performance over a broad state space. Implementation in real hardware would also require precise stereo calibration and real-time computation of joint-space inverse kinematics, both of which may present engineering challenges.

Overall, this work demonstrates that combining classical control design with geometric insights and modern parameterization techniques can yield a robust, efficient, and scalable solution to a longstanding problem in visual servoing. Future research may explore multi-rate updates, gain-scheduled switching between linearized controllers, or simultaneous $H_\infty$ feedforward-feedback design [41] to further enhance robustness and applicability in unstructured environments.

In summary, this paper investigates the overdetermination problem in stereo-based IBVS tasks. While future improvements to the algorithm are possible, the proposed control policy has demonstrated significant potential as an accurate and fast solution for real-world eye-in-hand (EIH) visual servoing tasks.

## Appendix A

In this section, we will show the geometric model of a specific robot manipulator ABB IRB 4600 45/2.05[23] and a figure of a camera model: Zed 2 with dimensions [24]. This section also contains specification tables of robots' dimensions, camera, and motor installed inside the joints of manipulators.

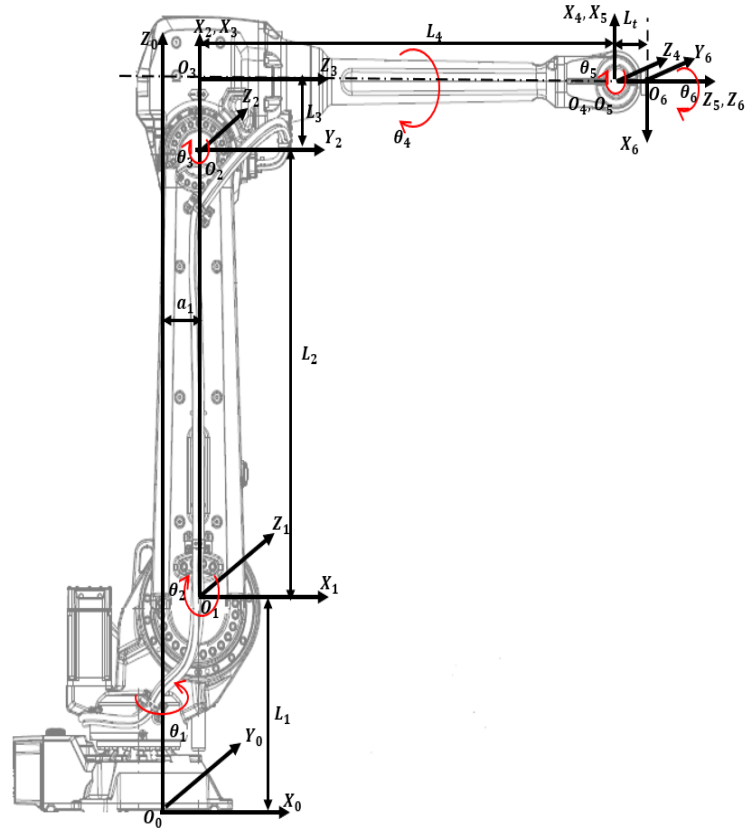

**Figure A1.** IRB ABB 4600 Model with attached frames.

*Dimensions are in mm*

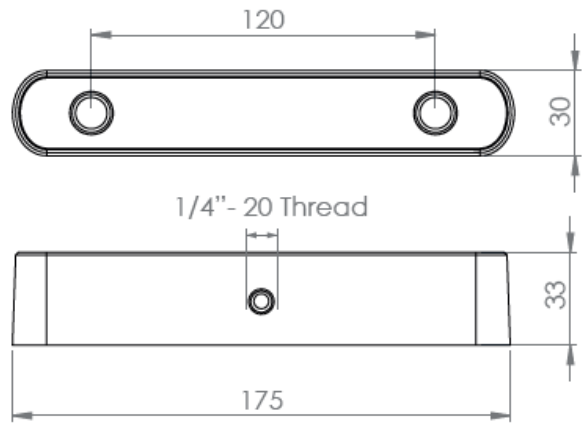

**Figure A2.** Zed 2 stereo camera model with dimensions.

**Table A1.** Specification Table of ABB IRB 4600 45/2.05 Model (Dimensions).

| Parameters | Values |
|---|---|
| Length of Link 1: $L_1$ | 495 mm |
| Length of Link 2: $L_2$ | 900 mm |
| Length of Link 3: $L_3$ | 175 mm |
| Length of Link 3: $L_4$ | 960 mm |
| Length of Link 1 offset: $a_1$ | 175 mm |
| Length of Spherical wrist: $L_t$ | 135 mm |
| Tool length (screwdriver): $\overline{PJ}_{tool}$ | 127 mm |

**Table A2.** Specification Table of ABB IRB 4600 45/2.05 Model (Axis Working range).

| Axis Movement | Working range |
|---|---|
| Axis 1 rotation | +180° to -180° |
| Axis 2 arm | +150° to -90° |
| Axis 3 arm | +75° to -180° |
| Axis 4 wrist | +400° to -400° |
| Axis 5 bend | +120° to -125° |
| Axis 6 turn | +400° to -400° |

**Table A3.** Specification Table of Stereo Camera Zed 2.

| Parameters | Values |
|---|---|
| Focus length: f | 2.8 mm |
| Baseline: B | 120 mm |
| Weight: W | 170g |
| Depth range: | 0.5m-25m |
| Diagonal Sensor Size: | 6mm |
| Sensor Format: | 16:9 |
| Sensor Size: W X H | 5.23mm X 2.94mm |
| Angle of view in width: $\alpha$ | 86.09° |
| Angle of view in height: $\beta$ | 55.35° |

**Table A4.** Specification Table of Motors and gears.

| Parameters | Values |
|---|---|
| **DC Motor** | |
| Armature Resistance: $R$ | 0.03 Ω |
| Armature Inductance: $L$ | 0.1 mH |
| Back emf Constant: $K_b$ | 7 mv/rpm |
| Torque Constant: $K_m$ | 0.0674 N/A |
| Armature Moment of Inertia: $J_a$ | 0.09847 kg$m^2$ |
| **Gear** | |
| Gear ratio: $r$ | 200:1 |
| Moment of Inertia: $J_g$ | 0.05 kg$m^2$ |
| Damping ratio: $B_m$ | 0.06 |

## Appendix B

In this section, we will show forward kinematics and inverse kinematics of the 6 DoFs revolute robot manipulators. The results are consistent with the model ABB IRB 4600.

Forward kinematics refers to the use of kinematic equations of a robot to compute the position of the end-effector from specified values for the joint angles and parameters. The equations are summarized in the below:

$$n_x = c_1 s_{23}(s_4 s_6 - c_4 c_5 c_6) - s_1(s_4 c_5 c_6 + c_4 s_6) - c_1 c_{23} s_5 c_6$$
$$n_y = s_1 s_{23}(s_4 s_6 - c_4 c_5 c_6) + c_1(s_4 c_5 c_6 + c_4 s_6) - s_1 c_{23} s_5 c_6$$
$$n_z = c_{23}(s_4 s_6 - c_4 c_5 c_6) + s_{23} s_5 c_6$$

(B1)

$$s_x = c_1 s_{23}(s_4 c_6 + c_4 c_5 c_6) + s_1(s_4 c_5 s_6 - c_4 c_6) + c_1 c_{23} s_5 s_6$$
$$s_y = s_1 s_{23}(s_4 c_6 + c_4 c_5 c_6) - c_1(s_4 c_5 s_6 - c_4 c_6) + s_1 c_{23} s_5 s_6$$
$$s_z = c_{23}(s_4 c_6 + c_4 c_5 c_6) - s_{23} s_5 s_6$$

(B2)

$$a_x = -c_1 s_{23} c_4 s_5 - s_1 s_4 s_5 + c_1 c_{23} c_5$$
$$a_y = -s_1 s_{23} c_4 s_5 + c_1 s_4 s_5 + s_1 c_{23} c_5$$
$$a_z = c_{23} c_4 s_5 - s_{23} c_5$$

(B3)

$$d_x = L_t(-c_1 s_{23} c_4 s_5 - s_1 s_4 s_5 + c_1 c_{23} c_5) + c_1(L_3 s_{23} + L_2 s_2 + a_1)$$
$$d_y = L_t(-s_1 s_{23} c_4 s_5 + c_1 s_4 s_5 + s_1 c_{23} c_5) + s_1(L_3 s_{23} + L_2 s_2 + a_1)$$
$$d_z = L_t(c_{23} c_4 s_5 - s_{23} c_5) + L_3 c_{23} + L_2 c_2 + L_1$$

(B4)

$$\text{Note: } c_i \equiv \cos(q_i), \ s_i \equiv \sin(q_i)$$
$$c_{i,j} \equiv \cos(q_i + q_j), \ s_{i,j} \equiv \sin(q_i + q_j)$$
$$i, j \in \{1,2,3,4,5,6\}$$

(B5)

where $[n_x, n_y, n_z]^T$, $[s_x, s_y, s_z]^T$ and $[a_x, a_y, a_z]^T$ are the end-effector's directional vector of Yaw, Pitch and Roll in base frame $O_0 X_0 Y_0 Z_0$ (Figure A1). And $[d_x, d_y, d_z]^T$ are the vector of absolute position of the center of the end-effector in base frame $O_0 X_0 Y_0 Z_0$. For a specific model ABB IRB 4600-45/2.05 (Handling capacity: 45 kg/ Reach 2.05m) the dimensions and mass are summarized in Table A1.

Inverse kinematics refers to the mathematical process of calculating the variable joint angles needed to place the end-effector in a given position and orientation relative to the inertial base frame. The equations are summarized in the below:

$$p_x = d_x - L_t a_x$$
$$p_y = d_y - L_t a_y$$
$$p_z = d_z - L_t a_z$$

(B6)

$$q_1 = arctan(\frac{p_y}{p_x})$$

(B7)

$$q_2 = \frac{pi}{2} - \arccos\left(\frac{L_2{}^2 + \left(\sqrt{p_x{}^2 + p_y{}^2} - a_1\right)^2 + (p_z - L_1)^2 - L_3{}^2 - L_4{}^2}{2 L_2 \sqrt{L_3{}^2 + L_4{}^2}}\right)$$
$$- arctan\left(\frac{p_z - L_1}{\sqrt{p_x{}^2 + p_y{}^2} - a_1}\right)$$

(B8)

$$q_3 = \pi - \arccos\left(\frac{L_2{}^2 + L_3{}^2 + L_4{}^2 - \left(\sqrt{p_x{}^2 + p_y{}^2} - a_1\right)^2 - (p_z - L_1)^2}{2 L_2 \sqrt{L_3{}^2 + L_4{}^2}}\right) - \arctan\left(\frac{L_4}{L_3}\right)$$

(B9)

$$q_5 = \arccos(c_1 c_{23} a_x + s_1 c_{23} a_y - s_{23} a_z)$$

(B10)

$$q_4 = \arctan\left(\frac{s_1 a_x - c_1 a_y}{c_1 s_{23} a_x + s_1 s_{23} a_y + c_{23} a_z}\right)$$

(B11)

$$q_6 = -\arctan\left(\frac{c_1 c_{23} s_x + s_1 c_{23} s_y - s_{23} s_z}{c_1 c_{23} n_x + s_1 c_{23} n_y - s_{23} n_z}\right)$$

(B12)

$$\text{Note: } c_i \equiv \cos(q_i), \ s_i \equiv \sin(q_i)$$
$$c_{i,j} \equiv \cos(q_i + q_j), \ s_{i,j} \equiv \sin(q_i + q_j)$$

(B13)

$$i, j \in \{1, 2, 3, 4, 5, 6\}$$

where $[n_x, n_y, n_z]^T$, $[s_x, s_y, s_z]^T$, $[a_x, a_y, a_z]^T$ and $[d_x, d_y, d_z]^T$ have been defined above in the forward kinematic discussion.

## Appendix C

In this section, we derive the transformation matrix $T_M^C$, representing the pose of the marker relative to the camera. This matrix is essential for hand-eye calibration, as defined in Equations (18) and (20).

Section 3 established that matching at least three 3D points is sufficient to uniquely determine the pose of a stereo camera system. Accordingly, three fiducial markers—denoted $R_1, R_2$ and $R_3$—are placed in the workspace. The centers of these markers are observed in the stereo camera system, with image coordinates given as:

$$p_{image,R_1} = [u_{l,R_1}, u_{r,R_1}, v_{R_1}]^T, \; p_{image,R_2} = [u_{l,R_2}, u_{r,R_2}, v_{R_2}]^T,$$
$$p_{image,R_3} = [u_{l,R_3}, u_{r,R_3}, v_{R_3}]^T. \tag{C1}$$

Equation (11) provides a projection function $M$ from 3D camera coordinates to 2D stereo image coordinates. Here, we define its inverse $M^{-1}$ which maps stereo image coordinates to 3D points in the camera frame:

$$\text{Stereo-camera mapping } M^{-1}: \; p_{image} = [u_l, u_r, v]^T \to P^C = [X^C, Y^C, Z^C]^T \tag{C2}$$

Using Equations (9)–(10), the relationships between 3D coordinates and image coordinates in a rectified stereo setup are:

$$Z^C \cdot u_l = F \cdot (X^C - b/2) \tag{C3}$$

$$Z^C \cdot u_r = F \cdot (X^C + b/2) \tag{C4}$$

$$Z^C \cdot v = F \cdot Y^C \tag{C5}$$

Solving Equations (C3)–(C5) yields the inverse stereo mapping:

$$X^C = \frac{b(u_r + u_l)}{2(u_r - u_l)} \tag{C6}$$

$$Y^C = \frac{b \cdot v}{(u_r - u_l)} \tag{C7}$$

$$Z^C = \frac{F \cdot b}{u_r - u_l} \tag{C8}$$

Equations (C6)–(C8) implement the inverse mapping $M^{-1}$ in Equation (C2), allowing us to convert each marker's observed image coordinates into 3D positions in the camera frame: $P_{R_1}^C$, $P_{R_2}^C$ and $P_{R_3}^C$.

We now define a marker coordinate frame, illustrated in Figure C1. The origin is placed at the center of one of the markers. The Z axis is defined to point perpendicular to the marker plane, facing toward the camera. The X axis points along the vector from the origin to another marker's center, and the Y axis is determined using the right-hand rule to complete the orthonormal frame. In this frame, the 3D positions of the marker centers are: $P_{R_1}^M$, $P_{R_2}^M$ and $P_{R_3}^M$.

Let $P_R^M = [P_{R_1}^M, P_{R_2}^M, P_{R_3}^M]$, and $P_R^C = [P_{R_1}^C, P_{R_2}^C, P_{R_3}^C]$. The transformation $T_M^C \in SE(3)$ maps marker coordinates to camera coordinates:

$$P_R^C = R \cdot P_R^M + t \tag{C8}$$

$$T_M^C = \begin{bmatrix} R & t \\ 0 & 1 \end{bmatrix} \tag{C9}$$

Here, $R \in \mathbb{R}^{3\times3}$ is the rotation matrix and $t \in \mathbb{R}^3$ is a translation vector.

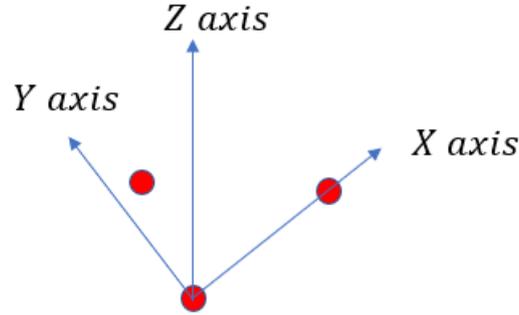

**Figure C1.** Marker's Coordinates frame.

We begin by computing the centroids of the point sets:

$$\overline{P_R^M} = \frac{1}{3}\sum_{i=1}^{3} P_{R_i}^M \tag{C10}$$

$$\overline{P_R^C} = \frac{1}{3}\sum_{i=1}^{3} P_{R_i}^C \tag{C11}$$

Then, we center the coordinates:

$$P_{R_i}^{M\prime} = P_{R_i}^M - \overline{P_R^M}, \quad i = 1,2,3 \tag{C12}$$

$$P_{R_i}^{C\prime} = P_{R_i}^C - \overline{P_R^C}, \quad i = 1,2,3 \tag{C13}$$

Compute the cross-covariance matrix:

$$H \in \mathbb{R}^{3\times3} = \sum_{i=1}^{3} P_{R_i}^{M\prime} \cdot (P_{R_i}^{C\prime})^T \tag{C14}$$

Apply Singular Value Decomposition (SVD):

$$H = U\Sigma V \tag{C15}$$

Here $U \in \mathbb{R}^{3\times3}$ is the left orthogonal matrix, $\Sigma \in \mathbb{R}^{3\times3}$ is the diagonal matrix with singular values, and $V \in \mathbb{R}^{3\times3}$ is the right orthogonal matrix.
The optimal rotation is:

$$R = VU^T \tag{C16}$$

And from Equation (C9), the optimal translation is:

$$t = \overline{P_R^C} - R \cdot \overline{P_R^M} \tag{C17}$$

Finally, the full transformation from marker to camera frame is:

$$T_M^C = \begin{bmatrix} VU^T & \overline{P_R^C} - VU^T \cdot \overline{P_R^M} \\ 0 & 1 \end{bmatrix} \tag{C18}$$